\documentclass{article}

\usepackage{microtype}
\usepackage{graphicx}
\usepackage{booktabs} 
\usepackage{multirow}
\usepackage{makecell}
\usepackage{arydshln}   
\usepackage{graphicx}
\usepackage{graphics}
\usepackage{bbm}
\usepackage{balance}
\usepackage{algorithm}
\usepackage{color}
\usepackage{amsmath}
\usepackage{amssymb}
\usepackage{textcomp}
\usepackage{color}
\usepackage[dvipsnames]{xcolor}
\usepackage{color}
\usepackage{subfig}

\usepackage{url}
\usepackage[accepted]{icml2024/icml2018}

\usepackage{array}
\usepackage{amsmath}
\usepackage{amssymb}
\usepackage{mathtools}
\usepackage{amsthm}
\usepackage{arydshln}
\usepackage[capitalize,noabbrev]{cleveref}
\usepackage{adjustbox} 

\theoremstyle{plain}
\newtheorem{theorem}{Theorem}[section]

\theoremstyle{definition}
\newtheorem{definition}[theorem]{Definition}

\theoremstyle{remark}

\usepackage[textsize=tiny]{todonotes}

\newcommand{\ignore}[1]{}

\icmltitlerunning{
\textsc{GliDe} with a \textsc{CaPE}:A Low-Hassle Method to Accelerate Speculative Decoding}

\begin{document}

\twocolumn[

\icmltitle{\adjustbox{valign=c}{\includegraphics[width=1.2cm]{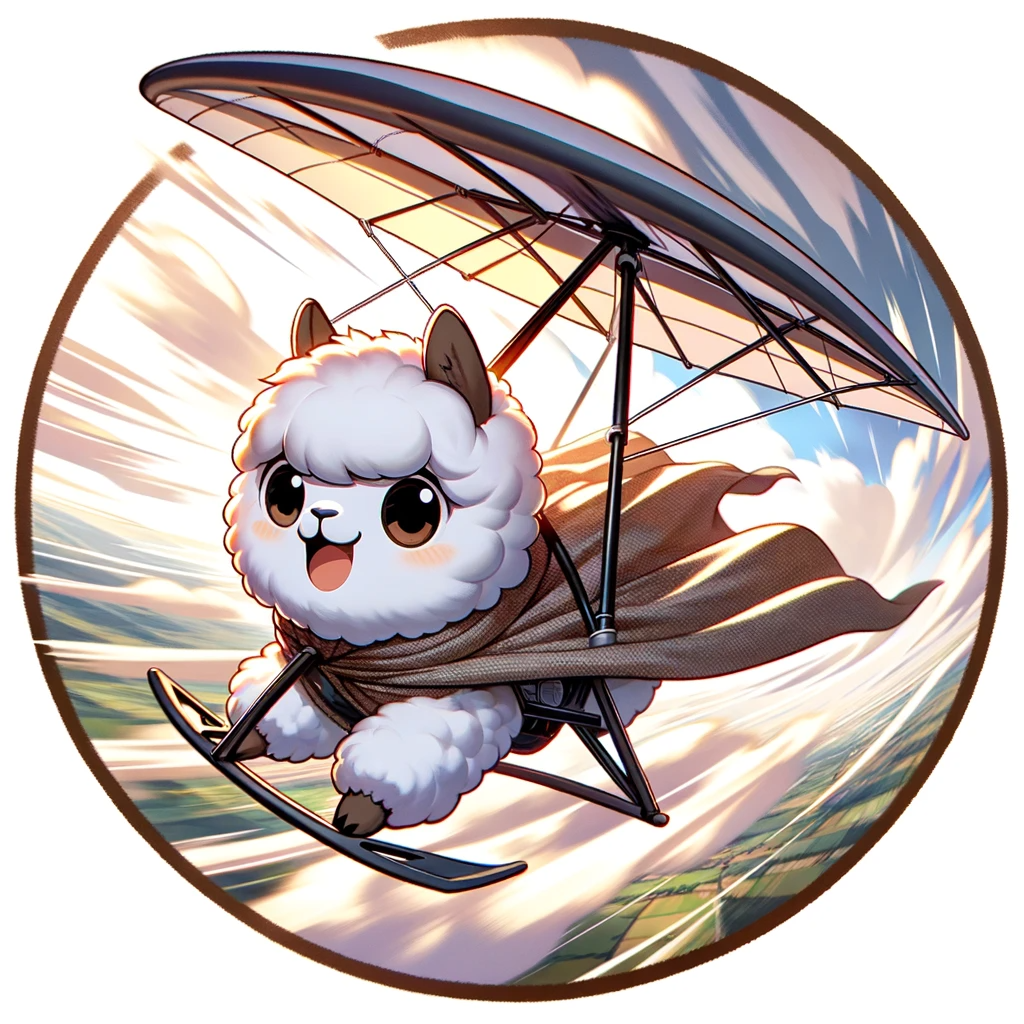}}~~\LARGE \textsc{GliDe} with a \textsc{CaPE}~~~~~~~~~~~~~~~}
\vspace{-1.7cm}
\icmltitle{A Low-Hassle Method to Accelerate Speculative Decoding}
\vspace{-0.5cm}


\icmlsetsymbol{equal}{*}

\begin{icmlauthorlist}
\icmlauthor{Cunxiao Du}{smu}
\icmlauthor{Jing Jiang}{smu}
\icmlauthor{Yuanchen Xu}{nus}
\icmlauthor{Jiawei Wu}{nus}
\icmlauthor{Sicheng Yu}{smu}
\icmlauthor{Yongqi Li}{polyu}
\icmlauthor{Shenggui Li}{nus}
\icmlauthor{Kai Xu}{nus}
\icmlauthor{Liqiang Nie}{hitsz}
\icmlauthor{Zhaopeng Tu}{tencent}
\icmlauthor{Yang You}{nus}
\end{icmlauthorlist}

\icmlaffiliation{smu}{Singapore Management University}
\icmlaffiliation{nus}{National University of Singapore}
\icmlaffiliation{polyu}{The Hong Kong Polytechnic University (PolyU)}
\icmlaffiliation{tencent}{Tencent AI Lab}
\icmlaffiliation{hitsz}{Harbin Institute of Technology (Shenzhen)}

\icmlcorrespondingauthor{Cunxiao Du}{cnsdunm@gmail.com}

\icmlkeywords{Large Language Model, Natural Language Process, Inference Acceleration}

\vskip 0.3in
]



\printAffiliationsAndNotice{} 

\begin{abstract}
Speculative decoding is a relatively new decoding framework that leverages small and efficient draft models to reduce the latency of LLMs.
In this study, we introduce \textsc{GliDe} and \textsc{CaPE}, two low-hassle modifications to vanilla speculative decoding to further improve the decoding speed of a frozen LLM.
Specifically, \textsc{GliDe} is a modified draft model architecture that reuses the cached keys and values from the target LLM, while \textsc{CaPE} is a proposal expansion method that uses the draft model's confidence scores to help select additional candidate tokens for verification. 
Extensive experiments on different benchmarks demonstrate that our proposed \textsc{GliDe} draft model significantly reduces the expected decoding latency. 
Additional evaluation using walltime reveals that \textsc{GliDe} can accelerate Vicuna models up to 2.17x and further extend the improvement to 2.61x with \textsc{CaPE}.
We will release our code, data, and the trained draft models.
\end{abstract}
\section{Introduction}
\label{sec:intro}

Large language models~(LLMs) have become increasingly powerful and are now adopted for a wide range of applications~\cite{jiao2023chatgpttranslation, llmecommerical}.
Many LLM applications require real-time responses, e.g., translation systems. 
However, LLMs are typically based on the autoregressive transformer architecture, which generates output tokens step by step and thus suffers from high latency, particularly with larger model sizes.
To reduce LLM serving latency, \emph{speculative decoding}~(SD) has been proposed as a viable solution~\cite{stern2018blockwise, deepmindChen2023AcceleratingLL, speculative-icml-2023}.
The key idea of SD is to use a smaller and more efficient \emph{draft} model to predict the next $\gamma$ tokens in the output sequence and then use the original LLM, which is called the \emph{target} model, to verify the proposed $\gamma$ tokens \emph{in parallel}.
SD hinges on the insight that not all output tokens are equally difficult to predict; employing a smaller but more efficient draft model that can correctly predict those ``easy'' tokens helps save inference time.
\begin{figure}[t]    \includegraphics[width=0.48\textwidth]{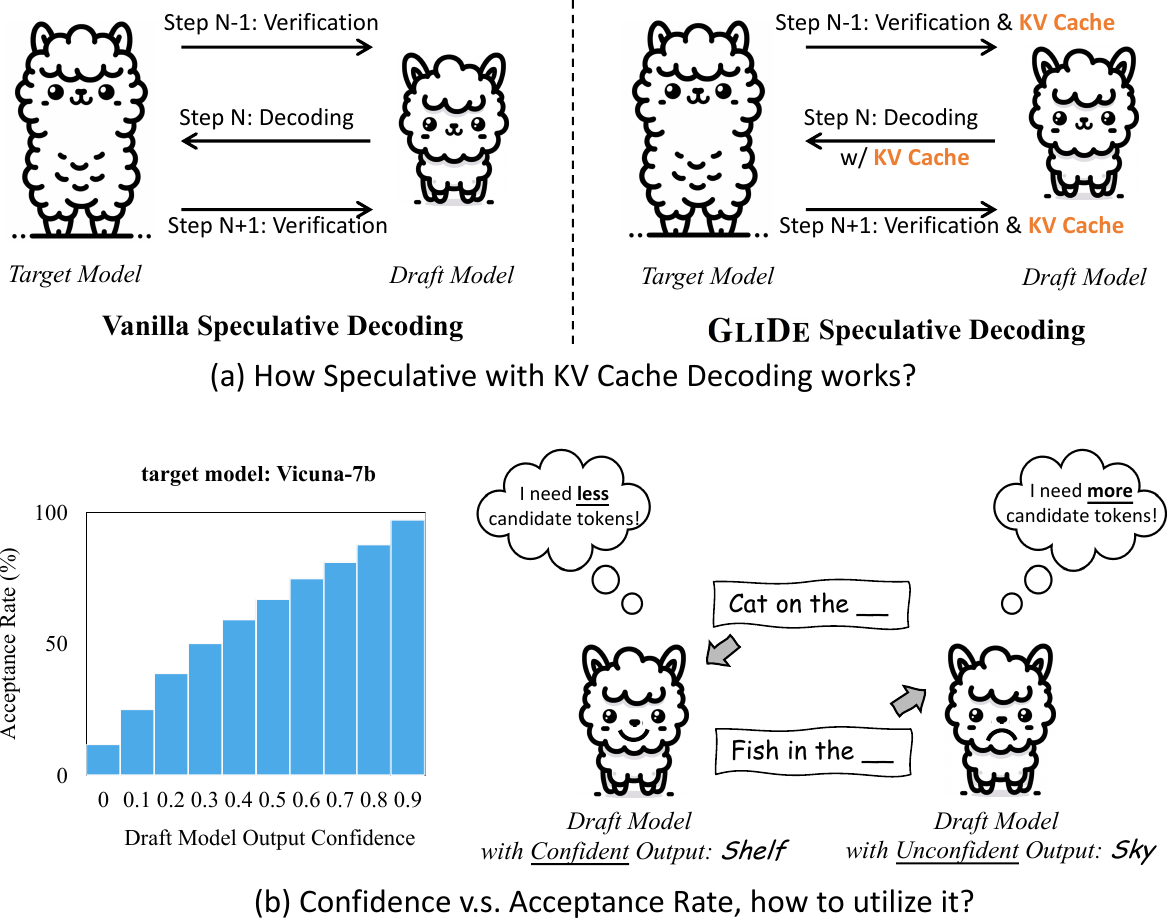}
    \caption{(a) Comparison between vanilla speculative decoding and our \textsc{GliDe}. (b) There is a positive correlation between acceptance rate and confidence score.}
    \label{fig:tisser}
\end{figure}
The success of SD relies heavily on the acceptance rate of the tokens proposed by the draft model, or in other words, how likely the target model will accept the proposed tokens.
Therefore, the design of the draft model and its proposal mechanism play a central role in accelerating SD.

Recent work on SD acceleration has explored two promising directions.
The first direction relies on the intuition that the more closely the draft model's predictions align with those of the target model, the greater the likelihood of the proposed tokens being accepted.
Along this line, researchers have proposed to align the draft model with the target model through distillation training~\cite{zhou2024distillspec, miao2023specinfer, liu2023onlinespeculativedecoding} or to use the same model for both speculation and verification~(i.e., self-speculative decoding)~\cite{fu2023lookaheaddecoding}.
Second, at each round of speculation, one can ask draft models to propose multiple candidate sequences for the target model to verify, thereby increasing the chance of acceptance.
In this vein, recent work explores employing multiple draft models to propose different candidate token sequences~\cite{miao2023specinfer} and using multiple prediction heads, one for each position, to propose multiple candidate token sequences in a non-autoregressive manner~\cite{medusa2023}.

In this work, we propose new solutions in the two aforementioned directions that are easy to implement and yet highly effective for accelerating SD.
Our solutions are motivated by the following insights.
First, we make an important observation that the draft model does not need to work separately from the target model during inference; by leveraging the KV~(key-value) cache of the target model, the draft model can propose tokens that are more likely to be accepted by the target model.
Second, we note that previous attempts to propose multiple candidate sequences are not ideal; they either rely on multiple draft models~\cite{miao2023specinfer}, which is not always feasible in practice, or use non-autoregressive decoding to keep the draft model's inference time low but sacrifice the fluency of the proposed candidate sequences~\cite{medusa2023}.
We propose a new solution that, without imposing much additional computational or memory requirements, simply expands a single candidate sequence with additional candidate tokens at each position.

Specifically, 
we introduce a cross-attention layer in the draft model's network architecture.
Although its implementation appears straightforward, this layer enables the draft model to access the target model's KV cache, a process that is empirically proven to be highly effective in our experiments.
We metaphorically describe this cross-attention as ``catching a glimpse'' and name our model \textsc{GliDe}, an acronym for \textbf{Gl}impse \textbf{D}raft Mod\textbf{e}l.
The idea is illustrated in Figure~\ref{fig:tisser}(a).
As we can see from the figure, because the target model stores its computed keys and values corresponding to the tokens it has verified in the last round of verification, these KV cache entries are free for the draft model to use.
By re-using these keys and values, the draft model is more likely to behave similarly to the target model.

To expand a proposed candidate sequence with additional candidate tokens, a key question is how many additional top-ranked tokens we should include at each position.
Although a naive solution is to use a pre-defined and fixed number, we suspect that we should provide more candidate tokens at a position where the top-1 candidate token has a low confidence score.
To verify whether this hypothesis holds, we first conduct a preliminary experiment to check whether a proposed token's prediction confidence is correlated with the chance of its acceptance.
Figure~\ref{fig:tisser}(b) shows the experiment results.
We can see that indeed there is a clear positive correlation between the draft model's prediction confidence of a proposed token and the token's chance of being accepted by the target model.
This observed correlation motivates us to design a \textbf{C}onfidence-\textbf{a}ware \textbf{P}roposal \textbf{E}xpansion~(\textsc{CaPE}) method that dynamically determines the number of additional candidate tokens to include in an expanded proposal sequence.

We conduct extensive experiments to evaluate the effectiveness of \textsc{GliDe} and \textsc{CaPE}.
Experiments on four datasets using Vicuna and Mistral as target models show that it is highly effective for \textsc{GliDe} to attend to the target model's KV cache, leading to an improvement of the acceptance rate of up to $23.5\%$, and \textsc{GliDe} significantly outperforms several baseline draft models, with an average improvement of $19.9\%$ in terms of acceptance rate, compared with previous draft models.
Additional experiments incorporating \textsc{CaPE} into \textsc{GliDe} reveal that \textsc{CaPE} can achieve speed increases ranging from 2.50x and 2.61x on different Vicuna models~\cite{Chiang2023Vicuna} based on walltime.









\section{Related Work}

Speculative decoding was first proposed as block-wise parallel decoding.
\citet{stern2018blockwise} trained multiple auxiliary models to predict $\gamma$ future tokens in parallel
and then used the original model to verify the future tokens in parallel.
Inspired by this idea, \citet{deepmindChen2023AcceleratingLL, biglittle, yuan2023speculativecontrastive, speculative-icml-2023} used an independent draft model to propose a short sequence of tokens for the target model to verify. There were also draft model-free works \cite{he2023rest, referencespeculative, fu2023lookaheaddecoding, selfspeculative} for speculative decoding.

\citet{speculative-icml-2023} pointed out that the main factors for speedup are the efficiency of the draft model and the acceptance rate of the proposed tokens. 
To improve the efficiency of the draft model,
\citet{medusa2023} proposed adding multiple LM Heads to predict future tokens at different positions independently.
To improve the acceptance rate, \citet{miao2023specinfer} proposed to generate multiple candidate sequences, which can be efficiently verified through a tree verification process~\cite{miao2023specinfer, medusa2023, sun2023spectrtree}.
Alternatively, distillation was used to train draft models that hopefully are similar to the target model.
\citet{zhou2024distillspec} proposed using sequence-level distillation~\cite{kim2016sequence} via the draft model's output,
while \citet{liu2023onlinespeculativedecoding} resorted to online distillation to quickly adapt the draft model to the current context. 

Our work focuses on the second factor, i.e., acceleration through improving acceptance rate.
\textsc{GliDe} re-uses the target model's KV cache to generate proposals more likely to be accepted, and \textsc{CaPE} expands the proposals with additional highly-ranked candidate tokens to further increase acceptance rate.

\section{Background: Speculative Decoding (SD)}

In vanilla SD, we assume that there is a powerful but slow LLM~(i.e., the \emph{target} model, denoted as $\mathcal{M}_\text{T}$) used to verify the final output sequence.
Meanwhile, a less powerful but faster language model~(i.e., the \emph{draft} model, denoted as $\mathcal{M}_\text{D}$) is used to propose candidate tokens.
Let $x_{\leq t}$ denote the prefix or prompt from which the next tokens are to be proposed.
SD works by first using the faster draft model $\mathcal{M}_\text{D}$ to autoregressively generate the next $\gamma$ tokens $(x_{t+1}, \ldots, x_{t+\gamma})$.
We refer to this proposed candidate token sequence as a \emph{proposal}.
However, because the proposal may not be what the target model $\mathcal{M}_\text{T}$ would have generated, the proposed tokens need to be \emph{verified} by $\mathcal{M}_\text{T}$.
$\mathcal{M}_\text{T}$ can verify the $\gamma$ tokens in parallel, which is much faster than autoregressive generation.
The verification step returns the first $n$ tokens ($0 \leq n \leq \gamma$) in the proposal that are accepted by $\mathcal{M}_\text{T}$ based on some acceptance strategy.
In addition, the verification step returns one more token $x'_{t+n+1}$ for free.
Then the sequence $(x_{t+1}, \ldots, x_{t+n}, x'_{t+n+1})$ is appended to the original prefix $x_{\leq t}$ to form the new prefix for the next round of speculation and verification.
There are two acceptance strategies: \emph{speculative decoding} and \emph{speculative sampling}. 
\citet{speculative-icml-2023} show that the outputs of these two strategies are equivalent to the outputs of the target model with greedy search and random sampling decoding strategies, respectively.




\section{Our Method: \textsc{GliDe} with a \textsc{CaPE}}
\label{sec:method}

As stated in $\S$\ref{sec:intro}, our method to accelerate SD consists of two parts:
(1) We design a draft model called \textsc{GliDe} that ``catches a glimpse of'' the KV cache computed during the \emph{target model's last round of verification} to assist the draft model's current round of proposal generation.
(2) We propose a proposal expansion mechanism called \textsc{CaPE} that uses the confidence scores of the draft model to dynamically determine how many additional top-ranked tokens to include at each position in a proposal sequence.

\subsection{\textsc{GliDe}: Glimpse Draft Model}

We assume that both the target and the draft models follow the standard decoder-only transformer architecture.
To enable the draft model to take advantage of the hidden representations of the prefix tokens computed and cached by the target model, we propose a new architecture for the draft model called \textsc{GliDe}~(Glimpse Draft Model).
\textsc{GliDe} allows the draft model $\mathcal{M}_\text{D}$ to re-use the cached key-value pairs in the target model $\mathcal{M}_\text{T}$, presumably making the distribution of $\mathcal{M}_\text{D}$ more consistent with that of $\mathcal{M}_\text{T}$ without incurring much additional computational cost.
Our $\mathcal{M}_\text{D}$ will be trained from scratch based on the \textsc{GliDe} architecture while $\mathcal{M}_\text{T}$ is kept frozen.

\ignore{
Let $N_{\text{T}}$ and $N_{\text{D}}$ denote the number of layers of $\mathcal{M}_\text{T}$ and $\mathcal{M}_\text{D}$, respectively.
Let $h$ denote the number of heads in each self-attention sub-layer of $\mathcal{M}_\text{T}$.
Let $d_k$ denote the dimension of the keys and values in $\mathcal{M}_\text{T}$.
Let $d_\text{D}$ denote the dimension of the embedding vectors in $\mathcal{M}_\text{D}$.
}

\begin{figure*}[h]
    \centering
    \includegraphics[width=0.85\textwidth]{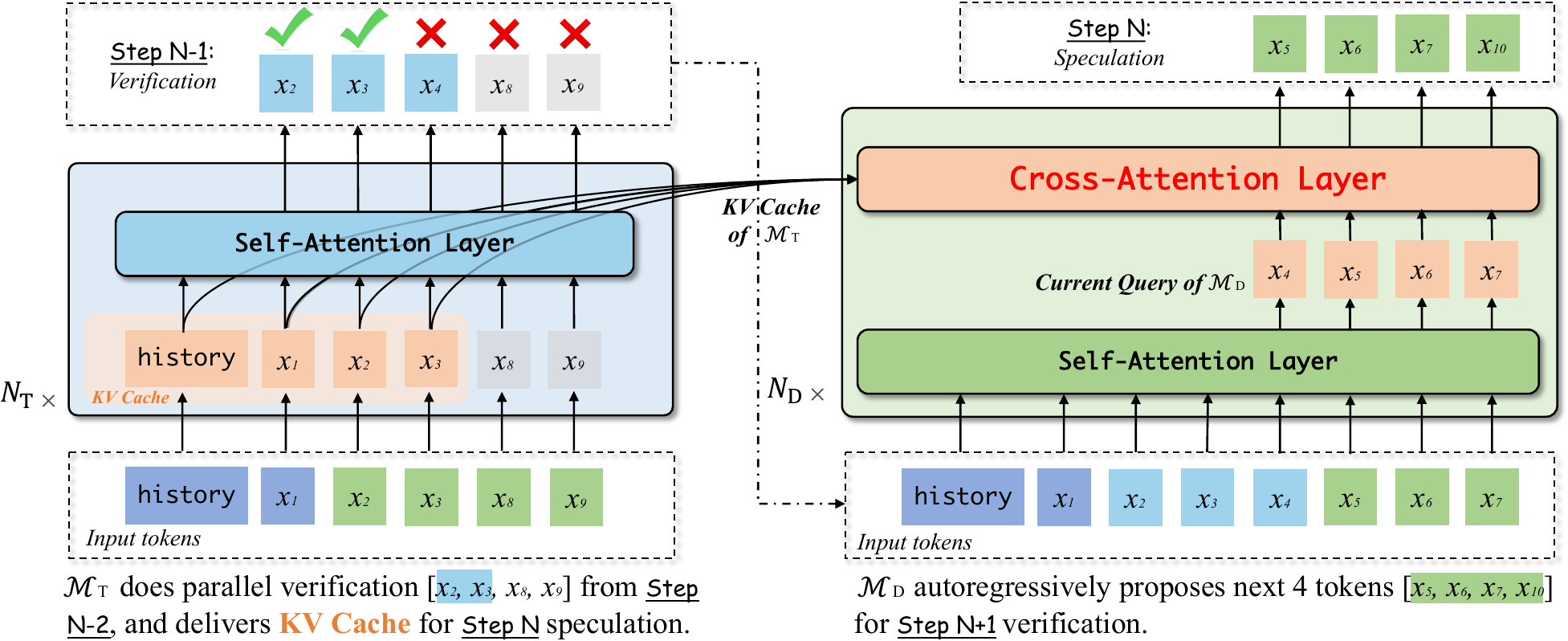}
    \caption{Overview of \textsc{GliDe}. We insert the \textcolor{orange}{\textbf{cross-attention}} layer in the Draft Model ($\mathcal{M}_{D}$) and let it glimpse the KV cache of the target model ($\mathcal{M}_{T}$) from the last verification.}
    \label{fig:model}
\end{figure*}

Specifically, assume that in the last round of speculation, $\mathcal{M}_\text{D}$ has proposed a sequence of tokens that is passed to $\mathcal{M}_\text{T}$ for verification.
Then in the subsequent verification step, assume that $\mathcal{M}_\text{T}$ accepts the sequence up to $x_{t-1}$ and generates an additional token $x_t$. 
After this verification step, we will keep only the KV cache for tokens up to position $(t-1)$ and discard the KV cache for those tokens rejected by $\mathcal{M}_\text{T}$.
The left hand side of Figure~\ref{fig:model} illustrates the verification process, where the grey cube represents the discarded KV cache. 

Now assume that given the prefix $x_{\leq t}$, the draft model $\mathcal{M}_\text{D}$ in its current round of speculation has proposed tokens $x_{t+1}$ to $x_{t+i-1}$.
Next, $\mathcal{M}_\text{D}$ is going to propose the next token $x_{t+i}$ based on the prefix $x_{<(t+i)}$.
In vanilla speculative decoding, $\mathcal{M}_\text{D}$ works independently of $\mathcal{M}_\text{T}$.
In our proposed \textsc{GliDe} architecture, however, we want to re-use $\mathcal{M}_\text{T}$'s cached keys and values associated with the prefix tokens up to $x_{t-1}$. 
\ignore{At this point, because $\mathcal{M}_\text{T}$ has just generated $x_t$, $\mathcal{M}_\text{T}$ has computed and stored the keys and values corresponding to all tokens up to and including $x_{t-1}$ at each of its self-attention sub-layer.}
To do so,
we insert a cross-attention sub-layer between the self-attention sub-layer and the feed-forward sub-layer in each transformer layer of $\mathcal{M}_\text{D}$, as shown in the right hand side of Figure~\ref{fig:model}.
This additional layer first projects the outputs between $t$ and $(t+i-1)$ from the self-attention sub-layer below into query vectors of a dimension that is compatible with $\mathcal{M}_\text{T}$'s keys and values.
Then this cross-attention sub-layer performs standard cross-attention between these queries from $\mathcal{M}_\text{D}$ and the KV cache from $\mathcal{M}_\text{T}$. The resulting vectors will be passed to the feed-forward sub-layer above.
 
Concretely, let $h$ denote the number of heads in each self-attention sub-layer of $\mathcal{M}_\text{T}$.
Let $(K^{l, j}, V^{l, j})$ denote the keys and values of the $j$-th head at the $l$-th layer of $\mathcal{M}_\text{T}$, where $K^{l, j}, V^{l, j} \in \mathbb{R}^{(t-1) \times d_k}$.
Let $H^m \in \mathbb{R}^{i \times d_\text{D}}$ represent the last $i$ output vectors from the $m$-th self-attention sub-layer of $\mathcal{M}_\text{D}$.
These $i$ vectors in $H^m$ correspond to positions $t$ to $(t+i-1)$.
We intend to use these vectors from  $\mathcal{M}_\text{D}$ as queries to attend to the representations of the prefix tokens up to $x_{t-1}$ in  $\mathcal{M}_\text{T}$.
Because $d_\text{D}$ (the dimension of the hidden vectors in $\mathcal{M}_\text{D}$) is generally different from $d_k$ (the dimension of the keys in $\mathcal{M}_\text{T}$), and because $\mathcal{M}_\text{T}$ has $h$ attention heads, we first perform $h$ linear projections to project $H^m$ into $h$ different query matrices, one for each head in $\mathcal{M}_\text{T}$:
\begin{eqnarray*}
Q^{m, j} & = & H^m W_j,
\end{eqnarray*}
where $j \in [1, h]$, and $W_j \in \mathbb{R}^{d_\text{D} \times d_k}$ are learnable parameters.
To perform cross-attention, each layer in $\mathcal{M}_\text{D}$ attends to a corresponding layer in $\mathcal{M}_\text{T}$, counting from the top layers.
In this way, we try to use the KV cache from the upper layers of $\mathcal{M}_\text{T}$ because presumably they are more contextualized and therefore better representations.
Specifically, 
let $N_{\text{T}}$ and $N_{\text{D}}$ denote the numbers of layers of $\mathcal{M}_\text{T}$ and $\mathcal{M}_\text{D}$, respectively.
For the $m$-th layer in $\mathcal{M}_\text{D}$, we set $l = N_\text{T} - N_\text{D} + m$ and let it attend to the $l$-th layer in $\mathcal{M}_\text{T}$ through multi-head attention as follows:
\begin{eqnarray*}
\text{MultiHead}(Q^m, K^l, V^l) & = & \text{Concat}(\text{hd}_1, \ldots, \text{hd}_h) W^O, \\
\text{where} \; \text{hd}_j & = & \text{SoftMax}(\frac{Q^{m, j} (K^{l, j})^{\intercal}}{\sqrt{d_k}}) V^{l, j}.
\end{eqnarray*}
Here $W^O \in \mathbb{R}^{ h d_k \times d_{\text{D}}}$ 
is a standard learnable parameter matrix that projects the concatenated heads back to dimension $d_\text{D}$.
$\text{MultiHead}(Q^m, K^l, V^l)$ will then be concatenated with the other output vectors (up to position $t$) from the $m$-th self-attention sub-layer of $\mathcal{M}_\text{D}$ and fed into the $m$-th feed-forward sub-layer of $\mathcal{M}_\text{D}$, as shown in Figure~\ref{fig:model}.

\paragraph{Block-wise Attention Mask.} 
We train the draft model $\mathcal{M}_\text{D}$ from scratch similar to the standard training of autoregressive decoder-only models.
A main difference is that during training, the target model $\mathcal{M}_\text{D}$ (which is kept frozen) is used to provide the KV cache for the cross-attention sub-layers of $\mathcal{M}_\text{D}$.
However, care must be taken to ensure consistency between the training stage and the inference stage regarding the use of $\mathcal{M}_\text{T}$'s KV cache.
Recall that at inference time when $\mathcal{M}_\text{D}$ is used for speculation, after $\mathcal{M}_\text{T}$ has verified tokens up to $x_{t-1}$ and generated token $x_t$, and when $\mathcal{M}_\text{D}$ is about to speculate the $(t+i)$-th token, $\mathcal{M}_\text{D}$ only has access to the KV cache in $\mathcal{M}_\text{D}$ up to position $(t-1)$ rather than position $(t+i-1)$.
This can be considered a KV cache delay.
When training the draft model $\mathcal{M}_\text{D}$, we need to simulate this delayed KV cache to ensure that the trained $\mathcal{M}_\text{D}$ works well with delayed KV cache.

To do so, we introduce a training mechanism with a block-wise attention mask as follows.
We divide the training sequences into blocks of length $L$~(which is set to be 5 in our experiments).
During training, when the draft model $\mathcal{M}_\text{D}$ is predicting token $x_j$ that is in the $i$-th block of a sequence, in $\mathcal{M}_\text{D}$'s cross-attention sub-layer, we will use the representations of only the tokens in the $i$-th block to the left of $x_j$ as queries.
These queries will attend to the KV cache in $\mathcal{M}_\text{T}$ corresponding to the tokens in the first $(i-1)$ blocks of the sequence but not tokens in the $i$-th block.
Formally, the block-wise attention mask takes an attention matrix $A$ as input and is defined as follows:
\begin{equation*}
\begin{aligned}
\text{Mask}_\text{Block}(A_{jk}) &= 
\begin{cases} 
A_{jk}, & \text{if } \text{block}(j) > \text{block}(k) \\
-\infty, & \text{otherwise}
\end{cases},
\end{aligned}
\end{equation*}
where $\text{block}(j)$ is a function that returns the index of the block where token $x_j$ belongs.
This attention mask is only used for the cross-attention sub-layers of the draft model during training and is not used for testing.

\subsection{\textsc{CaPE}: Confidence-Aware Proposal Expansion}

\begin{figure}[t]
    \centering
    \includegraphics[width=0.48\textwidth]{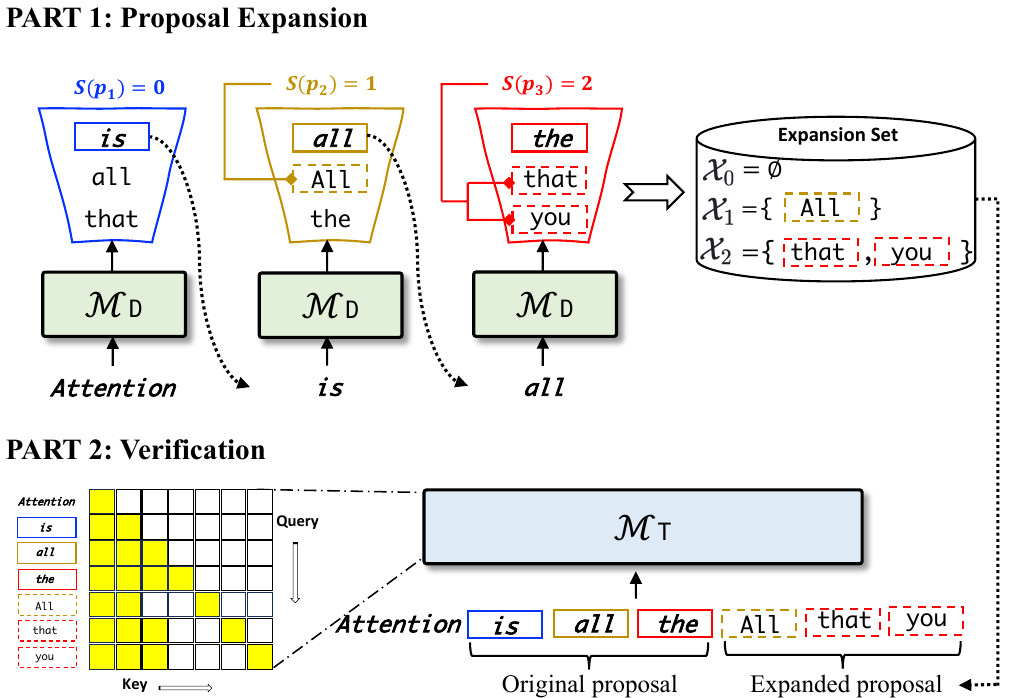}
    \caption{Overview of \textsc{CaPE}.}
    \label{fig:cape}
\end{figure}

Our \textsc{CaPE} has two components: a proposal expansion mechanism and a corresponding verification mechanism.
Figure~\ref{fig:cape} gives an overview of \textsc{CaPE}.

\paragraph{Proposal Expansion.} 
A standard proposal generated by a draft model given prompt $x_{\le t}$ consists of a sequence of $\gamma$ tokens, which we denote as $(x_{t+1}, x_{t+2}, \ldots, x_{t + \gamma})$.
These tokens have the highest probabilities at their corresponding positions as predicted by the autoregressive draft model.
In \textsc{CaPE}, we introduce a representation called \emph{expanded proposal} that augments a standard proposal with a sequence of expansion sets, one for each token in the proposal.
The expansion set $\mathcal{X}_i$ for $x_{t+i}$ in the proposal contains other top-ranked tokens at position $(t+i)$.
Formally, an expanded proposal is a sequence of tokens $(x_{t+1}, x_{t+2}, \ldots, x_{t+\gamma})$ where each token $x_{t+i}$ is associated with an expansion set $\mathcal{X}_i = \{x'_{t+i, j}\}_{j=1}^{K_i}$ with each $x'_{t+i, j}$ being a token and $K_i$ being the size of $\mathcal{X}_i$.

\ignore{
\begin{definition}[Expanded Proposal]
An expanded proposal is a sequence of tokens $(x_{t+1}, x_{t+2}, \ldots, x_{t+\gamma})$ where each token $x_{t+i}$ is associated with an expansion set $\mathcal{X}_i = \{x'_{t+i, j}\}_{j=1}^{K_i}$ with each $x'_{t+i, j}$ being a token and $K_i$ being the size of $\mathcal{X}_i$.
\end{definition}
}

With minor modifications to the standard proposal generation procedure, an expanded proposal is constructed as follows.
Let $p_\text{D}(\cdot | x_{<t+i})$ denote the next token distribution computed by the draft model $\mathcal{M}_\text{D}$.
Similar to standard proposal generation, the token that has the highest probability according to $p_\text{D}(\cdot | x_{<t+i})$ is selected to be $x_{t+i}$ and appended to the proposal.
Let $p_i = \max p_\text{D}(\cdot | x_{<t+i})$ denote the probability (which can be regarded as a confidence score) of predicting this $x_{t+i}$.
A function $S(\cdot)$ is then used to determine the size $K_i$ of the expansion set $\mathcal{X}_i$ based on the confidence score $p_i$.
As discussed in $\S$\ref{sec:intro}, when the confidence $p_i$ is high, we expect the top-ranked $x_{t+i}$ to be correct with a high chance and therefore will use a small $K_i$.
When $p_i$ is low, we will use a relatively large $K_i$.
Concretely, we set $S(p)$ to be 7, 5, 3, and 1 for $p$ in the ranges of $(0, 0.3]$, $(0.3, 0.6]$, $(0.6, 0.8]$, and $(0.8, 1]$, respectively.
Once $K_i$ is determined, we will select $K_i$ tokens (excluding $x_{t+i}$) that have the highest probabilities according to $p_\text{D}(\cdot | x_{<t+i})$ and place them in $\mathcal{X}_i$.
These tokens are the next best choices for position $(t+i)$ other than $x_{t+i}$.

It is important to point out that identification of the highest-ranked tokens to be placed in $\mathcal{X}_i$ can be done while the draft model continues to autoregressively generate the following tokens beyond $x_{t+i}$.
Therefore, the construction of the expansion sets does not slow down the draft model's inference.
It is also important to note that tokens in the expansion sets are not used for predicting future tokens.
This is a key difference from proposal expansion using beam search, where each of the top-$k$ tokens at a position is used to generate subsequent tokens, resulting in expanded proposals in tree structures with many branches.
Such beam search-based expansion would be much more expensive to compute and thus less efficient, which we will demonstrate in $\S$\ref{sec:exp}.

\ignore{
Recall that we use the confidence score $p_i$ to determine the size of $\mathcal{X}_i$.
This is motivated by our observation that those proposed tokens with less certainty~(i.e., lower confidence scores) are more likely to be rejected by the target model.
Figure~\ref{fig:tisser} (b) shows our preliminary experiments testing how the acceptance rate changes for proposed tokens with different confidence scores.
We can see that clearly those tokens with high confidence are more likely to be accepted.
Based on this observation, we define the function $S(\cdot)$ as follows so that the size of the expansion set is larger for tokens with lower confidence.
The function $S(p)$ we use in our experiments is equal to 7, 5, 3, and 1 for the $p$ in the range of $(0, 0.3]$, $(0.3, 0.6]$, $(0.6, 0.8]$ and $(0.8, 1]$.
}

 
\paragraph{Verification of Expanded Proposals.}
Because our proposals now contain additional candidate tokens at each position, the verification mechanism also needs to be modified.
We borrow ideas from the token tree verifier~\cite{miao2023specinfer, medusa2023, sun2023spectrtree} and implement our verification procedure as follows.
First, we \emph{linearize} an expanded proposal into a single sequence by simply appending the tokens in the expansion sets to the end of the proposal. 
We then pass this sequence to $\mathcal{M}_{T}$ for parallel verification with a special causal mask.
The key idea of this mask is to ensure that each token, regardless of whether it is in the original proposal or in an expansion set, attends to only those tokens in the original proposal in front of this token and the token itself.

Mathematically, the mask is defined as follows. 
Let $\mathcal{P} = (x_{t+1}, \ldots, x_{t+\beta})$ denote the linearized expanded proposal, where the first $\gamma$ tokens are from the original proposal and the remaining tokens are from the expansion sets.
Let $\text{pos}(i)$ be a function that maps the token $x_{t+i}$ in $\mathcal{P}$ to the token's original position.
That means, for $i > \gamma$, $\text{pos}(i) = j$ where $x_{t+i} \in \mathcal{X}_j$.
Let $A$ denote a $\beta \times \beta$ matrix representing the attention scores between the tokens in $\mathcal{P}$.
The mask function is defined as:
\begin{equation*}
\begin{aligned}
\text{mask}_{\textsc{CaPE}}(A)_{ij} & = 
\begin{cases} 
A_{ij} & \text{if }  j \leq \gamma \text{ and } \text{pos}(i) > j \\
A_{ij} & \text{if }  i = j \\
-\infty & \text{otherwise } 
\end{cases}.
\end{aligned}
\end{equation*}
Here the condition $j \leq \gamma$ is to check whether $x_{t+j}$ is a token from the original proposal, the condition $\text{pos}(i) > j$ is to ensure that token $x_{t+i}$ 
is after token $x_j$, and the condition $i = j$ is for self-attention. 

\section{Experiments}
\label{sec:exp}

\begin{table*}[t]
\centering
\setlength{\tabcolsep}{5pt}
\begin{tabular}{lr cc cc cc cc}
\toprule
\multirow{2}{*}{\bf ~~~~~~~~~Models} & \multirow{2}{*}{\bf Cost} & \multicolumn{2}{c}{\textbf{GSM8K}} & \multicolumn{2}{c}{\textbf{Fin.-Alp.}} & \multicolumn{2}{c}{\textbf{Spider}} & \multicolumn{2}{c}{\textbf{Code.}} \\
 \cmidrule(lr){3-4}\cmidrule(lr){5-6} \cmidrule(lr){7-8} \cmidrule(lr){9-10}
 &$\downarrow(\%)$ &Acc. \footnotesize{(\%)} &$\mathbb{E}$(Spd.) 
&Acc. \footnotesize{(\%)} &$\mathbb{E}$(Spd.)
&Acc. \footnotesize{(\%)}&$\mathbb{E}$(Spd.)
&Acc. \footnotesize{(\%)}&$\mathbb{E}$(Spd.)\\
\midrule
 $\mathcal{M}_{T}$~\textbf{Vicuna-7B}\\\cmidrule{0-0}
~~~~~~~~~\footnotesize{\textsc{LLaMA}-68m} &7.7 &51.6 &1.46 &49.7 &1.41 &19.2 &0.89 &34.8 &1.11\\
\footnotesize{$\mathcal{M}_{D}$~~\textsc{LLaMA}-160m} &49.0 &58.0 &0.66 &57.1 &0.65 &25.1 &0.38 &40.0 &0.48 \\
~~~~~~~~~\footnotesize{\textsc{GliDe}-47m} & \bf 6.7  &~~\textbf{64.8}$^{\dagger}$ &~~\textbf{1.97}$^{\dagger}$ &~~\textbf{63.2}$^{\dagger}$ &~~\textbf{1.90}$^{\dagger}$ &~~\textbf{55.7}$^{\dagger}$ &~~\textbf{1.64}$^{\dagger}$ &~~\textbf{67.0}$^{\dagger}$&~~\textbf{2.06}$^{\dagger}$ \\
\midrule
 $\mathcal{M}_{T}$~\textbf{Vicuna-13B}\\\cmidrule{0-0}
~~~~~~~~~\footnotesize{\textsc{LLaMA}-68m} &6.2 &50.2 &1.51 &49.3 &1.48 &19.0 &0.94 &32.7 &1.13\\
\footnotesize{$\mathcal{M}_{D}$~~\textsc{LLaMA}-160m} &39.3 &56.8 &0.75 &56.8 &0.75 &25.4 &0.45 &38.3 &0.54 \\
~~~~~~~~~\footnotesize{\textsc{GliDe}-47m} & \bf 5.5 &~~\textbf{67.1}$^{\dagger}$ &~~\textbf{2.16}$^{\dagger}$ &~~\textbf{65.0}$^{\dagger}$ &~~\textbf{2.07}$^{\dagger}$ &~~\textbf{57.1}$^{\dagger}$ &~~\textbf{1.77}$^{\dagger}$ &~~\textbf{68.6}$^{\dagger}$ 
&~~\textbf{2.24}$^{\dagger}$  \\
\midrule
 $\mathcal{M}_{T}$~\textbf{Vicuna-33B}\\\cmidrule{0-0}
~~~~~~~~~\footnotesize{\textsc{LLaMA}-68m} &\textbf{4.6} &49.1 &1.57 &46.5 &1.50 &30.0 &1.16 &31.5 &1.19\\
\footnotesize{$\mathcal{M}_{D}$~~\textsc{LLaMA}-160m} &29.5 &56.1 &0.89 &53.1 &0.84 &31.4 &0.59 &36.7 &0.64 \\
~~~~~~~~~\footnotesize{\textsc{GliDe}-71m} & \bf 6.8 
&~~\textbf{69.3}$^{\dagger}$  &~~\textbf{2.16}$^{\dagger}$  
&~~\textbf{64.3}$^{\dagger}$  &~~\textbf{1.94}$^{\dagger}$  
&~~\textbf{62.7}$^{\dagger}$  &~~\textbf{1.87}$^{\dagger}$  
&~~\textbf{68.9}$^{\dagger}$ &~~\textbf{2.14}$^{\dagger}$  \\
\midrule
 \multicolumn{4}{l}{$\mathcal{M}_{T}$~\textbf{Mistral-7B-Ins.}}\\\cmidrule{0-0}
\footnotesize{$\mathcal{M}_{D}$~~}\footnotesize{\textsc{LLaMA}-45m}$^{*}$ &\textbf{4.9} &36.7 &1.27 &40.2 &1.34 &41.9 &1.38 &44.5 &1.44 \\
~~~~~~~~~\footnotesize{\textsc{GliDe}-47m} &  \bf 6.6 &~~\textbf{60.1}$^{\dagger}$  &~~\textbf{1.80}$^{\dagger}$  &~~\textbf{56.4}$^{\dagger}$  &~~\textbf{1.67}$^{\dagger}$  &~~\textbf{59.8}$^{\dagger}$  &~~\textbf{1.79}$^{\dagger}$  &~~\textbf{62.6}$^{\dagger}$  &~~\textbf{1.89}$^{\dagger}$ \\
\bottomrule
\end{tabular}
\caption{Comparison between \textsc{GliDe} and previous draft models.
$^\dagger$ denotes results that are statistically significantly better than the corresponding best \textsc{LLaMA} draft model with $p<0.01$.
*Since there is no open-source draft model trained for mistral, we retrained it using the same data of \textsc{GliDe} following the setting of~\citet{miao2023specinfer}. }
\label{tab:main}
\end{table*}

\subsection{Settings}


\paragraph{Target and draft models.} 
We select two widely-used LLMs, Vicuna~(including 7b, 13b, and 33b)~\cite{Chiang2023Vicuna} and Mistral~(7b-instruct-v0.1)~\cite{Jiang2023Mistral7}, as target models.
To make the draft model more efficient, we choose a wider and shallower architecture.
Specifically, for 7b and 13b target models, we use set $N_\text{D}$ to 1, and for the 33b target model, we set $N_\text{D}$ to 2.
We set $d_\text{D}$ to be $4096$.
Standard cross entropy is used to optimize the draft model while the parameters of the target model are kept frozen.
More details on model training can be found in Appendix~\ref{apd: details}.

\paragraph{Datasets.} 
We first train our draft model on the pre-training dataset SlimPajama-6B~\cite{cerebras2023slimpajama}. 
We then finetune the draft model on
a supervised-finetuning~(SFT) dataset (ShareGPT~\cite{sharegpt} in our case) to further improve the model performance.
Following~\citet{liu2023onlinespeculativedecoding}, we evaluate our \textsc{GliDe} method across four different datasets: GSM8K~\cite{gsm8k}~(math reasoning), Finance-Alpaca~\cite{bharti2023financealpaca}~(QA for finance), Spider~\cite{yu2018spider}~(text-to-SQL), and Code-Search-Python~\cite{codesearchnet}~(Python code generation). 
We follow~\cite{medusa2023} and use the well-known benchmark dataset MT-Bench~\cite{mtbench} for the evaluation of \textsc{CaPE}.

\paragraph{Metrics.}
A widely-used metric for SD is \emph{acceptance rate} $\alpha$~\cite{speculative-icml-2023},
which is the expected probability of the target model accepting a token proposed by the draft model.
\citet{speculative-icml-2023} showed that with speculative sampling, $\alpha$ is equivalent to $\mathbb{E}_{x \sim p_\text{D}(x)}\min(p_\text{T}, p_\text{D})$, where $p_\text{T}$ and $p_\text{D}$ are the target and the draft models' next token probabilities.
The acceptance rate is independent of hardware configuration and therefore a more objective metric.

\citet{speculative-icml-2023} defined the \emph{cost coefficient} $c$ to be the ratio between the walltime of a single run of $\mathcal{M}_\text{D}$ and that of $\mathcal{M}_\text{T}$.
Given proposal length $\gamma$, acceptance rate $\alpha$, and cost coefficient $c$, \citet{speculative-icml-2023} derived the following formula for the expected improvement factor in total walltime, or \emph{expected speedup}.
\begin{equation*}
\begin{aligned}
\mathbb{E}(\text{Spd.}) = \frac{1 - \alpha^{\gamma+1}}{(1 -\alpha)(\gamma c + 1)}.
\end{aligned}
\end{equation*}

Finally, we also use actual decoding speed and walltime speedup for additional comparison.

\paragraph{Experiment design.}

To verify the effectiveness of \textsc{GliDe} and \textsc{CaPE} separately, we design two sets of experiments.
First,
we employ only \textsc{GliDe} without \textsc{CaPE} and use acceptance rate and expected speedup as evaluation metrics.
We set proposal length $\gamma$ to be 5 and adopt speculative sampling as our acceptance strategy, following~\cite{liu2023onlinespeculativedecoding}.
Next, we compare \textsc{GliDe}+\textsc{CaPE} with Medusa~\cite{medusa2023} and \textsc{GliDe}+BeamSearch.
Both Medusa and BeamSearch generate multiple proposals and use a tree verification mechanism.
Following Medusa~\cite{medusa2023}, here we adopt speculative decoding as our acceptance strategy, and the batch size is set to be 1.





\subsection{Evaluation of \textsc{GliDe}}


\paragraph{Effectiveness of attending to target model's KV cache.}

\begin{figure}[t]
    \centering \includegraphics[width=0.45\textwidth]{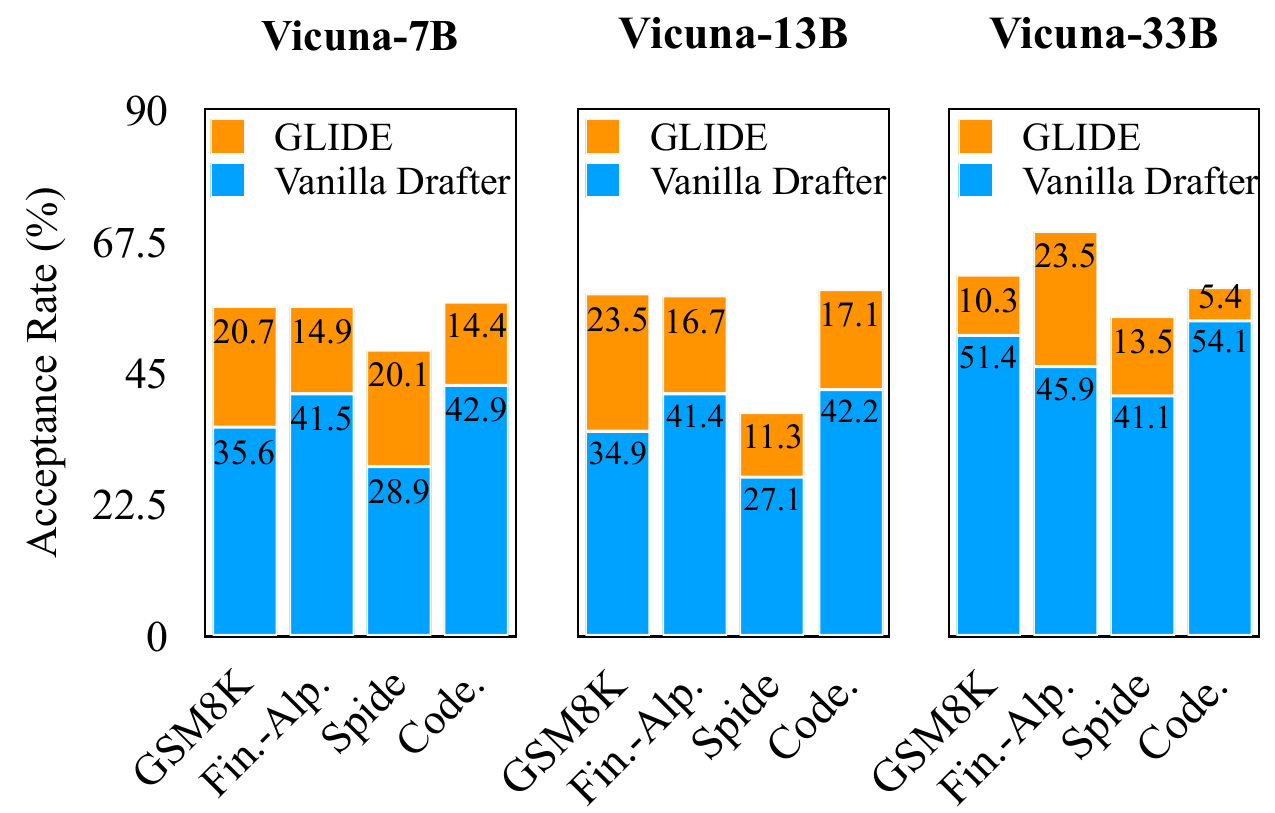}
    \caption{Comparison between \textsc{GliDe} and Vanilla Drafter. The oranges boxes show the \emph{improvement} brought by \textsc{GliDe}.}
    \label{fig: ablation}
    \vspace{-0.4cm}
\end{figure}

To verify whether it is effective to use the target model's KV cache,
we first conduct a controlled experiment where we compare two draft model architectures: (1) \textsc{GliDe}, which uses cross-attention to re-use the target model's KV cache, and (2) Vanilla Drafter, which has the same architecture as \textsc{GliDe} except that the cross-attention sub-layers are removed.
To reduce the experimental cost, we randomly sample half of the data from SlimPajma-6B to train both \textsc{GliDe} and the Vanilla Drafter.
We further fine-tune the two models on ShareGPT.
We run the experiments on GSM8K, Fin.-Alp., Spider, and Code.
We find that \textsc{GliDe} always has a substantially higher acceptance rate than Vanilla Drafter.
In Figure~\ref{fig: ablation}, we show the improvement in terms of acceptance rate by \textsc{GliDe} over Vanilla Drafter on the four datasets.
We can see that the improvement of acceptance rate ranges between 5.4 and 23.5 percentage points, and in most settings the improvement is over 10 percentage points.
Given such substantial improvement, we can conclude that it is highly effective for the draft model to re-use the target model's KV cache to improve the quality of the proposed sequences and thus improve the acceptance rate.

\paragraph{Comparison between \textsc{GliDe} and other draft models.}

Next, we compare our \textsc{GliDe} draft model trained on the entire set of SlimPajma-6B and finetuned on ShareGPT with a few previous baseline draft models, namely, \textsc{LLaMA}-68m~\cite{miao2023specinfer}, \textsc{LLaMA}-160m~\cite{miao2023specinfer}, and \textsc{LLaMA}-45m.
Comparison between our \textsc{GliDe} and these baseline draft models is shown in Table~\ref{tab:main}.
We show three metrics: Cost~(cost coefficient), Acc.~(acceptance rate), and $\mathbb{E}(\text{Spd.})$~(expected speedup).

From the table, we have the following findings.
(1) The cost coefficient of \textsc{GliDe} is very low, comparable to that of \textsc{LLaMA}-68m or \textsc{LLaMA}-45m.
Recall that the cost coefficient measures the relative walltime cost of the draft model compared with the target model, and a 5\%-6\% cost coefficient means the walltime cost of our \textsc{GliDe} draft model is negligible compared with that of the target model.
(2) In terms of acceptance rate and expected speedup, our \textsc{GliDe} draft model clearly beats the baseline draft models under all settings, and the improvement is all statistically significant. 
(3) The acceptance rate of \textsc{GliDe} generally falls between 55\% and 70\%, which means more than half of the proposed tokens are accepted.
In comparison, the baseline draft models sometimes only have an acceptance rate of around 30\%.
Similarly, the expected speedup of \textsc{GliDe} ranges between 1.67 and 2.24, much higher than those of the baseline draft models, which is always below 1.5 for \textsc{LLaMA}-68m and \textsc{LLaMA}-45m and below 1.0 for \textsc{LLaMA}-160m.
In sum, the results in Table~\ref{tab:main} again demonstrate that \textsc{GliDe} 
is highly effective.

\begin{figure}[t]
    \centering \includegraphics[width=0.5\textwidth]{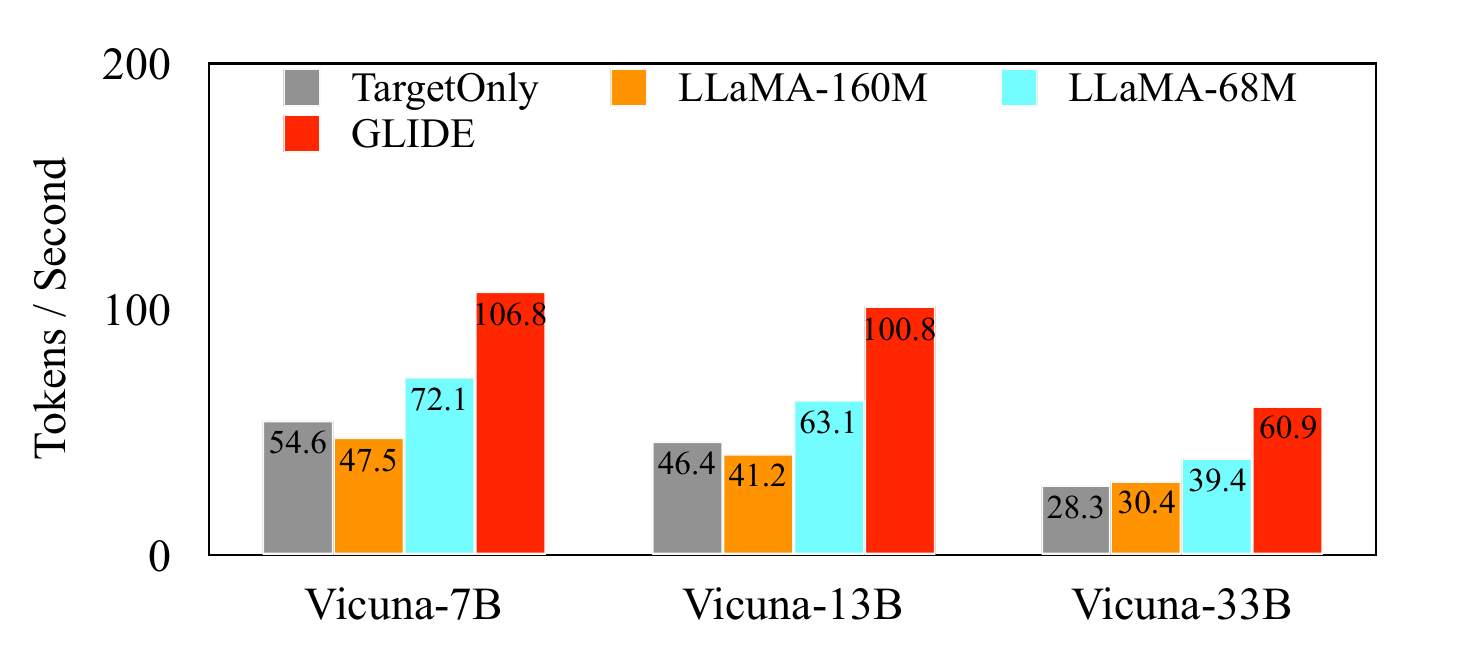}
    \vspace{-0.6cm}
    \caption{The decoding speed (tokens per second) on MT-Bench.}
    \label{fig: mtbench}
    \vspace{-0.5cm}
\end{figure}

\paragraph{Comparison of actual decoding speed.}
To see the actual speedup of \textsc{GliDe}, we also report the actual decoding speed in terms of number of tokens per second based on 
MT-bench~\cite{mtbench}.
The comparison between the speed of \textsc{GliDe} and that of the baselines is shown in Figure~\ref{fig: mtbench}. 
We find that \textsc{GliDe} substantially accelerates the model's decoding speed.
The results are consistent with the findings with the expected speedup. 
It is interesting to note that when using \textsc{GliDe}, our accelerated vicuna-33b model has a faster decoding speed than vicuna-7b without speculative decoding.

\subsection{Evaluation of \textsc{CaPE}}

\begin{figure}[t]
\hspace{0.25cm}
    \centering \includegraphics[width=0.5\textwidth]{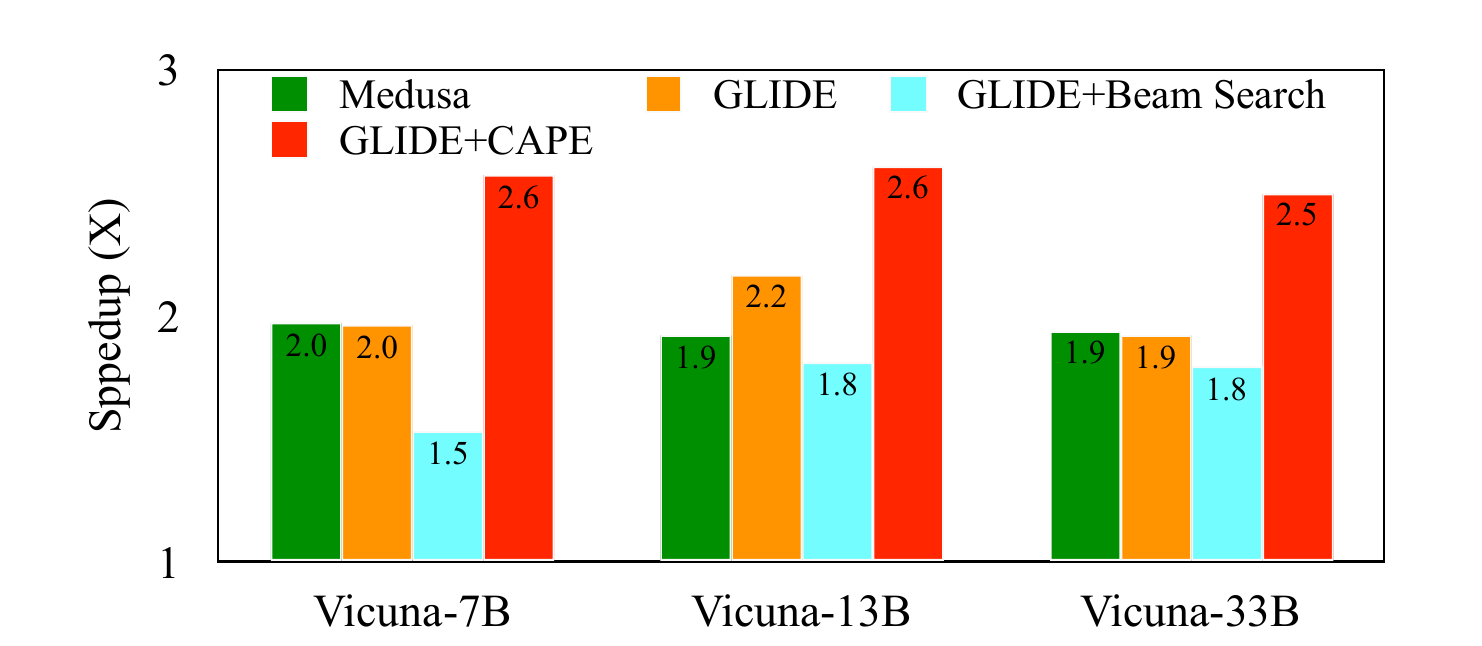}
    \vspace{-0.6cm}
    \caption{The practical speedup with \textsc{GliDe} + \textsc{CaPE}.}
    \label{fig: speedup}
    \vspace{-0.5cm}
\end{figure}

Next, we assess the effectiveness of \textsc{CaPE} on top of \textsc{GliDe}.
It is worth noting that acceptance rate based on its original definition is not applicable to the setting when the draft model proposes additional candidate tokens for verification.
Therefore, 
in this set of experiments, we directly measure the walltime speedup on MT-Bench, in line with Medusa.
Here, the walltime speedup is defined to be the ratio between the walltime of applying the speculative decoding method under evaluation and the walltime of standard decoding using only the target model.
We compare \textsc{GliDe}+\textsc{CaPE} with the following baselines:
(1) Medusa~\cite{medusa2023}, which uses a non-autoregressive draft model to parallel generate multiple proposals.
(2) \textsc{GliDe}, which does not use \textsc{CaPE}.
(3) \textsc{GliDe}+BeamSearch, where we set the beam size equal to 4.
Medusa and \textsc{GliDe}+BeamSearch use a tree attention mechanism~\cite{miao2023specinfer, medusa2023} for efficient verification.
The results are shown in Figure~\ref{fig: speedup}.

We have the following findings from the figure.
(1) \textsc{GliDe}+\textsc{CaPE} clearly outperforms all the baselines with a substantial margin. 
(2) There is a clear improvement of walltime speedup when \textsc{CaPE} is added on top of \textsc{GliDe}.
This illustrates the usefulness of further employing \textsc{CaPE} on top of \textsc{GliDe}.
(3) It is interesting to observe that even without \textsc{CaPE}, \textsc{GliDe} alone performs better than Medusa.
This again shows the effectiveness of \textsc{GliDe} itself.
It is worth pointing out that Medusa also uses the hidden states from the target model for its draft model's proposal generation, but instead of using all the previous KV cache like what we do, Medusa uses the the hidden state of the last verified tokens only, which may affect the quality of the draft model's predictions.
Furthermore, Medusa uses a non-autoregressive way to generate proposal sequences.
It is well known that non-autoregressive language models tend to generate output sequences with lower fluency~\cite{NAT, Du2021OAXE}.
In comparison, our \textsc{GliDe} model is an autoregressive model.
Therefore, the quality of the proposals by Medusa is likely lower than ours.
We suspect that this is another important factor for Medusa to perform worse than our method in terms of walltime speedup.
(4) Another interesting finding is that \textsc{GliDe}+BeamSearch is slower than \textsc{GliDe}.
This shows that simply employing beam search to generate multiple proposal sequences is not guaranteed to work.
Although the additional proposal sequences may increase the chance of acceptance, computationally, generating these proposals during speculation and verifying them during verification incur additional costs, which in our experiments seem to outweigh their benefits.
We conduct further analysis in $\S$\ref{subsec:analysis}.

It is worth noting that our \textsc{GliDe}+\textsc{CaPE} method is not only faster than Medusa but also verifies a smaller number of proposed tokens in each batch than Medusa.
Our \textsc{CaPE} sets the maximum number of tokens for verification at each step to be 32, whereas Medusa's is 64. 
So our \textsc{CaPE} may support larger batch size inference.

\subsection{Further Analysis}
\label{subsec:analysis}

\begin{figure}[t]
    \centering
    
    \includegraphics[width=0.75\linewidth]{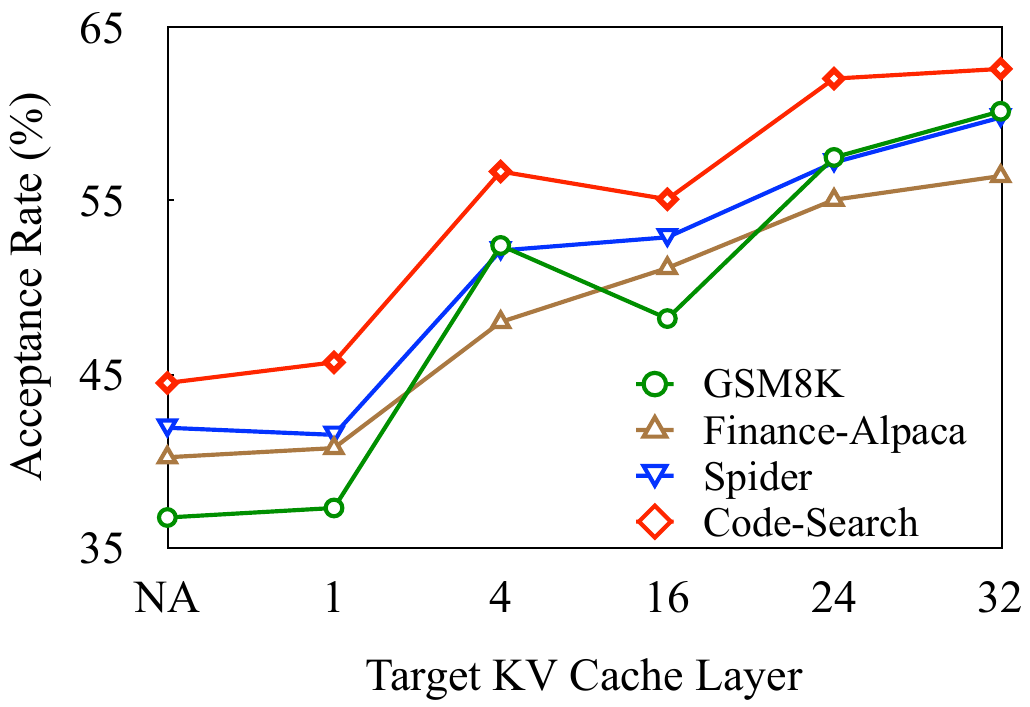}
    \vspace{-0.3cm}
    \caption{The relationship between the acceptance rate and the $n$-th layer of target model for KV cache.} 
    \label{fig: largekvab}
    \vspace{-0.5cm}
\end{figure}

\paragraph{Impact of KV cache at different layers.}
In our default setting, the draft model attends to the KV cache from the top layers of the target model, as described in $\S$\ref{sec:method}.
It is also possible to attend to lower layers' KV cache.
To see whether indeed KV cache from higher layers is more effective, we compare the acceptance rates when \textsc{GliDe} is used with Mistral-7b as the target model on the four datasets.
The results are shown in Figure~\ref{fig: largekvab}.
We can see that clearly using KV cache from higher layers of the target model produces higher acceptance rates, confirming our assumption that using the KV cache from the top layers is more effective.
On the other hand, using KV cache from lower layers is still useful compared with not using KV cache at all.
Therefore, \textsc{GliDe} can potentially be combined with early exit speculative decoding methods~\cite{calm} to further reduce the overall decoding time.

\paragraph{\textsc{CaPE} vs. beam search.}
Earlier in Figure~\ref{fig: speedup} we find that while \textsc{CaPE} reduces walltime, beam search incurs more walltime.
To better understand the issue, we separately compare the speculation time and verification time.
We find that for verification, \textsc{CaPE} and beam search have similar walltime.
But for each step of speculation, beam search increases the walltime from 5.5ms to 10.9ms~(a difference of 5.4ms) on average, whereas \textsc{CaPE} only increases the walltime by 0.2ms.
The main reason for beam search to increase walltime so much is that beam search requires copying and sorting operations that are very time-consuming.
This phenomenon was also observed by other researchers\footnote{\url{https://github.com/ggerganov/llama.cpp/issues/3137}}.
\paragraph{The importance of confidence scores in \textsc{CaPE}.}
Recall that \textsc{CaPE} uses the draft model's confidence scores to dynamically determine the number of additional tokens in each expansion set.
To see whether this confidence-aware expansion is useful, we compare \textsc{CaPE} with another proposal expansion method, where we set the size of the expansion sets to 4, which is the average size of the practical expansion sets.
We find that using this fixed-size expansion method, the decoding speed decreases from 70.6 tokens/sec to 68.4 tokens/sec on MT-Bench.
This shows that it is useful to use the confidence scores to adjust the sizes of different expansion sets.

\section{Conclusions}

In this work, we propose a draft model architecture called \textsc{GliDe} that leverages the KV cache from the target model to improve its proposal generation.
We also propose a confidence-aware proposal expansion mechanism called \textsc{CaPE} that produces additional candidate tokens for verification.
Experiments demonstrate that both \textsc{GliDe} and \textsc{CaPE} are highly effective ways to accelerate speculative decoding.
Our method also substantially outperforms the strong baseline Medusa based on walltime.
Overall, the integration of \textsc{GliDe} with \textsc{CaPE} results in a 2.5x speedup on Vicuna models.
As future work, we will explore the batch serving of \textsc{GliDe} and its effectiveness in processing extremely long contexts.

\ignore{
introduce a novel draft language model named \textsc{GliDe}, designed to glimpse and leverage the information from a target large language model.
Additionally, we propose a Confidence-Aware Proposal Expansion (\textsc{CaPE}) mechanism, aimed at increasing \textsc{GliDe}'s speed when generating multiple output proposals.
Our approach has been validated across five renowned benchmarks, demonstrating that \textsc{GliDe} markedly improves the decoding speed compared to earlier draft models. 
We validated our approach on five well-known benchmarks and showed that \textsc{GliDe} significantly improves decoding speed over previous draft models.
The integration of \textsc{GliDe} with \textsc{CaPE} results in a further acceleration, exceeding a 2.5 times increase in speed on Vicuna models. 
}

\section{Impact Statement}
Our work significantly improves the inference speed at which large language models make predictions, thus helping to make AI technologies more accessible and opening up more possibilities for personal AI applications.
However, since our method does not change the output generated by LLMs, it means our approach could inadvertently speed up the creation of harmful or biased content, such as hate speech or misinformation. This highlights the critical need for careful use and the establishment of strong safeguarding measures to reduce the dangers linked to enhanced LLM processing speeds.

\bibliography{icml2024/ref}
\bibliographystyle{icml2024/icml2024}

\clearpage

\appendix

\section{Experiment Details}
\label{apd: details}
For the 7B and 13B target language models, we employ a single-layer \textsc{GliDe} with a hidden dimension of 4096. For the 33B target model, we use a two-layer \textsc{GliDe} also with a hidden dimension of 4096.

In the case of the 7B and 13B target models, we train \textsc{GliDe} with zero2 and eight H800 GPUs. For the 33B target model, we use zero3 and 16 H800 GPUs. We set batch size (with accumulation) as 64, learning rate equals to 5e-4, and use adamW~\cite{kingma2015adam} to optimize the draft model. We only train our draft model for one epoch on both pretrain and SFT datasets.

For the 7B and 13B models, the training is approximately 7 hours and 10 hours, respectively, whereas for the 33B target model, it takes about 100 hours. 

As discussed in the section limitations, the main training expense lies in forwarding the target model to get the KV cache. If we could incorporate \textsc{GliDe} during the training of the LLM, this time-consuming part could be omitted. 
All the inference processes in this paper are performed using fp16 and on a single H800 GPU.

\section{Impact of Distillation}
\label{apd: distill}

\begin{table}[t]
\centering
\begin{tabular}{l cccc}
\toprule

{\bf Models} &{\bf GSM8K} & {\bf Fin.-Alp.} & {\bf Spider} & {\bf Code-Py.} \\
\midrule
w/o Dist. &64.8 &63.2 &55.7 &67.0 \\
w Dist.  &65.0 &63.9 &56.0 &67.5 \\ 
\bottomrule

\end{tabular}
\caption{Impact of Distillation on \textsc{GliDe} w.r.t acceptance rate (\%).}
\label{apd tab: distill}
\end{table}
Distillation is a common technology for speculative decoding~\cite{zhou2024distillspec, liu2023onlinespeculativedecoding}. We use seq-level distillation~\cite{kim2016sequence} for \textsc{GliDe} to see the benefit.
Table~\ref{apd tab: distill} shows that our \textsc{GliDe} can be further improved via distillation. However, the training time cost of seq-level distillation is very expensive, which is 2 times than the training time of vanilla \textsc{GliDe}. So we do not use distillation in other experiments. 

\section{Walltime for Speculation}
\begin{table}[t]
\centering
\begin{tabular}{l cccc}
\toprule

{\bf Models} & {\bf \textsc{GliDe}} &{\bf Beam 4} & {\bf Beam 8} & {\bf +\textsc{CaPE}} \\
\midrule
Vicuna-7B &5.3 &10.9 &14.8 &5.5 \\
Vicuna-13B &5.3 &10.6 &14.9 &5.5 \\ 
Vicuna-33B &9.1 &17.1 &24.6 &8.9 \\ 
\bottomrule

\end{tabular}
\caption{The speed of draft model's speculation (ms).}
\label{apd tab: time comparsion}
\end{table}
We provide more walltime results at Table~\ref{apd tab: time comparsion}.

\section{Batch Serving}
\label{abd: batch}

Here we also show the decoding speed of our proposed \textsc{GliDe} at Table~\ref{tab: apd_batch}. Please note we do not design any specific algorithm for batch serving, so the experiment here is similar to the vanilla draft model.
To avoid other irrelevant reasons like different samples in the same batch having different lengths, we just duplicate the input IDs to batch numbers. We test the performance using one H800 and fp16 as the platform and setting.

\begin{table*}[t]
\centering
\begin{tabular}{l cc cc cc}
\toprule
\multirow{2}{*} {\bf Batch Size} & \multicolumn{2}{c}{\textbf{Vicuna-7B}} & \multicolumn{2}{c}{\textbf{Vicuna-13B}} & \multicolumn{2}{c}{\textbf{Vicuna-33B}} \\
 \cmidrule(lr){2-3}\cmidrule(lr){4-5} \cmidrule(lr){6-7} 
 
 &+\textsc{GliDe} &Only 
 &+\textsc{GliDe} &Only 
 &+\textsc{GliDe} &Only \\ 
\midrule
1 &106.8 &54.6 
& 100.8 & 46.2 
&60.9 &28.3 \\
2 &2~*~108.2 & 2~*~55.6 
& 2~*~99.5 & 2~*~ 44.5
& 2~*~59.8 & 2~*~ 27.9 \\
4 &4~*~116.5 &4~*~55.9 
& 4~*~100.8 & 4~*~42.9
& OOM & OOM \\
8 & 8~*~108.5 &8~*~55.1
& 8~*~98.5 & 8~*~ 43.2
& OOM & OOM \\
16 & 16~*~106.2 &16~*~55.4
& OOM & 16~*~43.2
& OOM & OOM \\
32 & OOM & OOM & OOM & OOM & OOM & OOM \\

\bottomrule
\end{tabular}
\caption{The speed (tokens/sec) of \textsc{GliDe} under different batch sizes settings. OOM denotes out of the HBM memory of 80GB H800.}
\label{tab: apd_batch}
\end{table*}
\begin{figure*}[h]
    \centering
    \begin{minipage}[b]{0.24\linewidth}
        \centering
        \includegraphics[width=\linewidth]{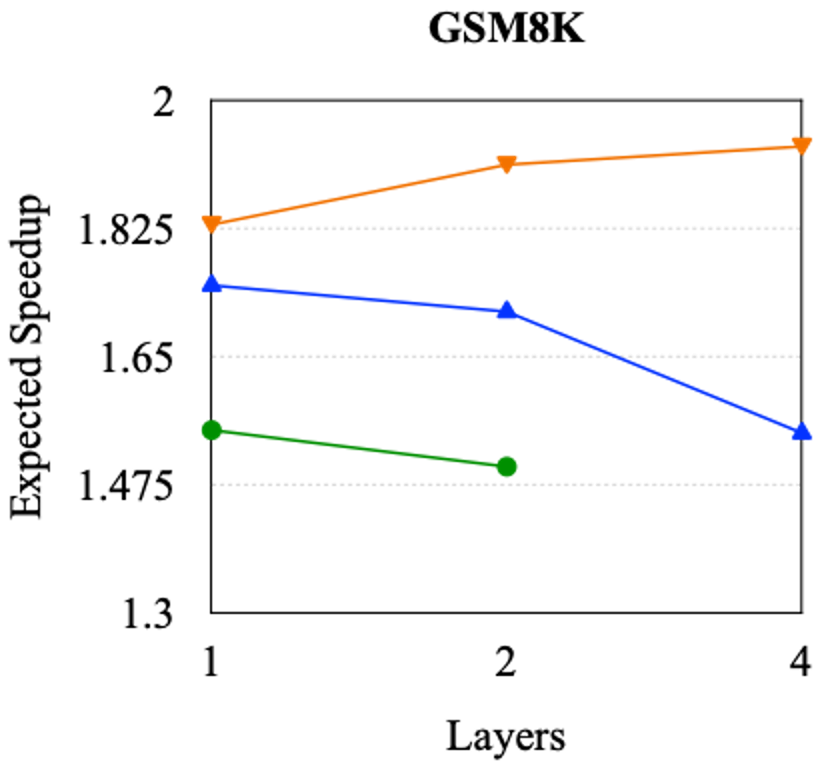}
        \label{fig:sub1}
    \end{minipage}
    \hfill 
    \begin{minipage}[b]{0.24\linewidth}
        \centering
        \includegraphics[width=\linewidth]{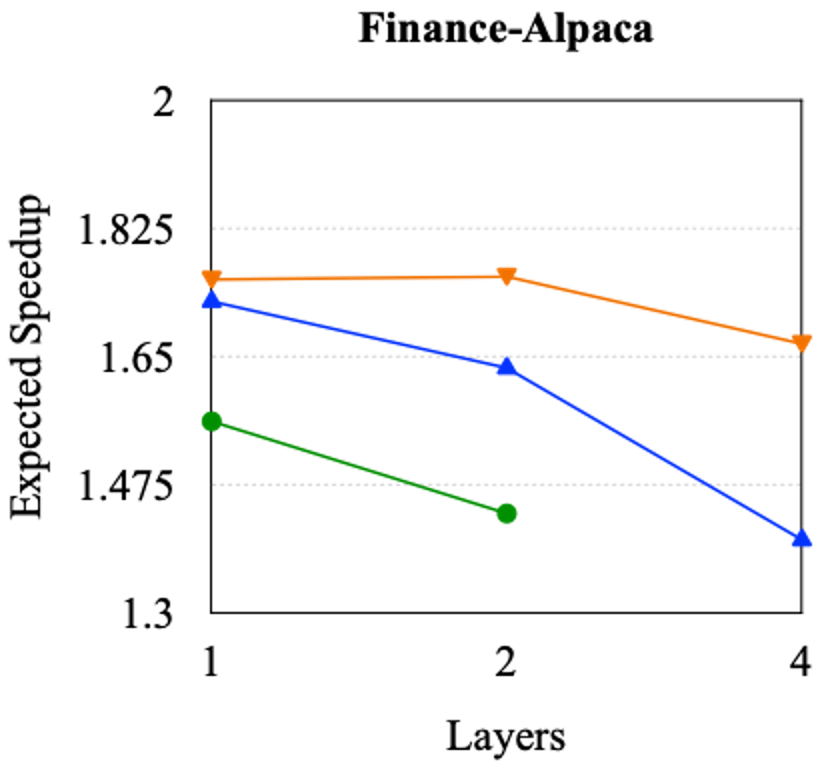}
        \label{fig:sub2}
    \end{minipage}
    \hfill
    \begin{minipage}[b]{0.24\linewidth}
        \centering
        \includegraphics[width=\linewidth]{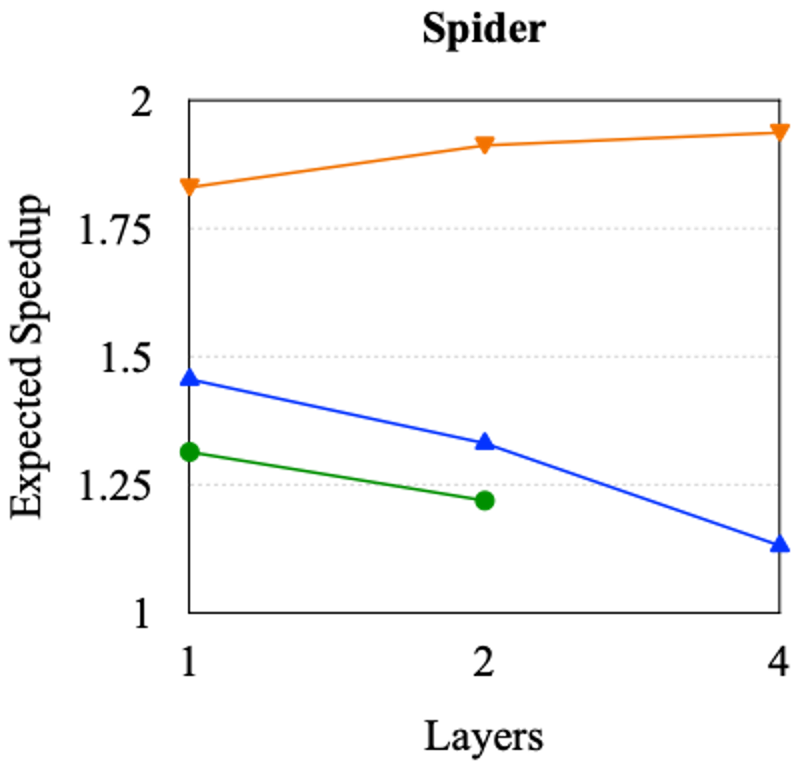}
        \label{fig:sub2}
    \end{minipage}
    \hfill
    \begin{minipage}[b]{0.24\linewidth}
        \centering
        \includegraphics[width=\linewidth]{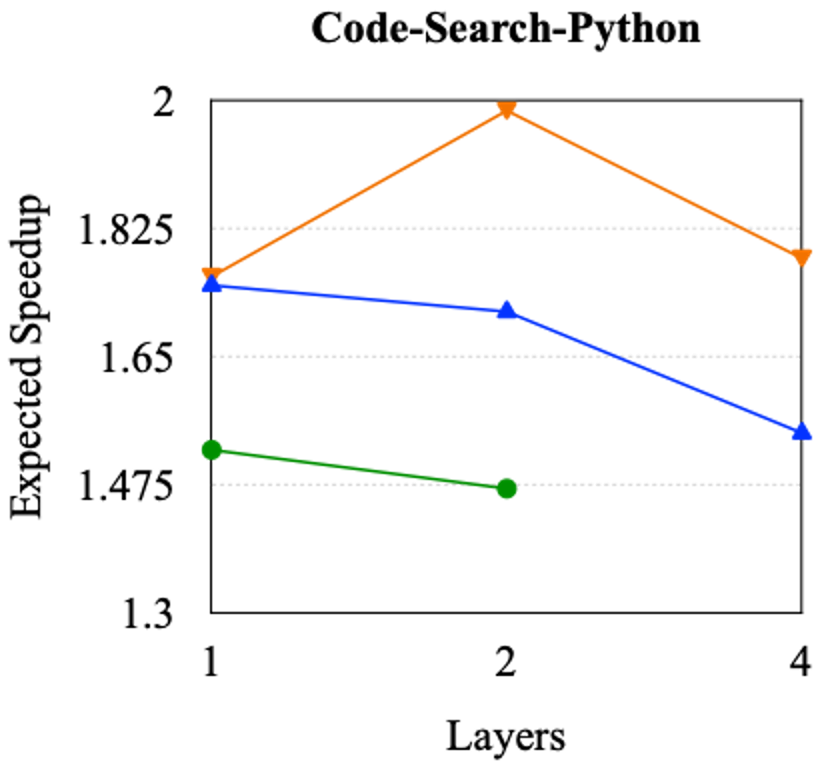}
        \label{fig:sub2}
    \end{minipage}
    \vspace{-0.6cm}
    \caption{The relationship between theoretical speedup and the number of layers in a draft model is such that when the target LLM has fewer layers, increasing the number of layers in the draft model results in a significant decrease in performance. This is because, compared to the improvement in the acceptance rate brought about by adding layers to the draft model, the increase in time cost is greater.}
    \label{fig: layer_scaling}
\end{figure*}

\begin{figure*}[!ht]
    \centering
    \begin{minipage}[b]{0.24\linewidth}
        \centering
        \includegraphics[width=\linewidth]{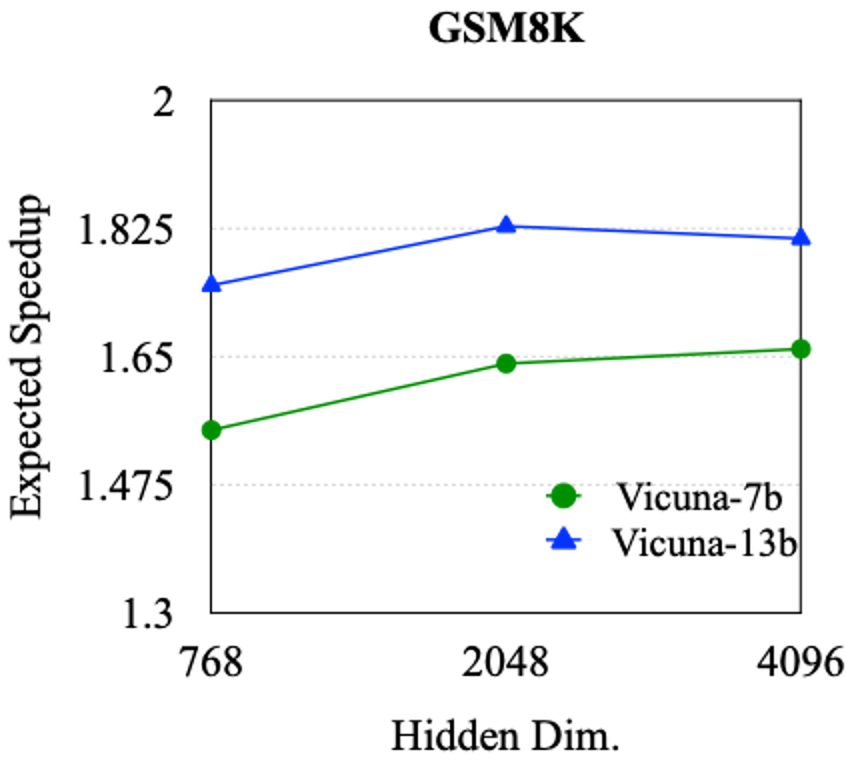}
        \label{fig:sub1}
    \end{minipage}
    \hfill 
    \begin{minipage}[b]{0.24\linewidth}
        \centering
        \includegraphics[width=\linewidth]{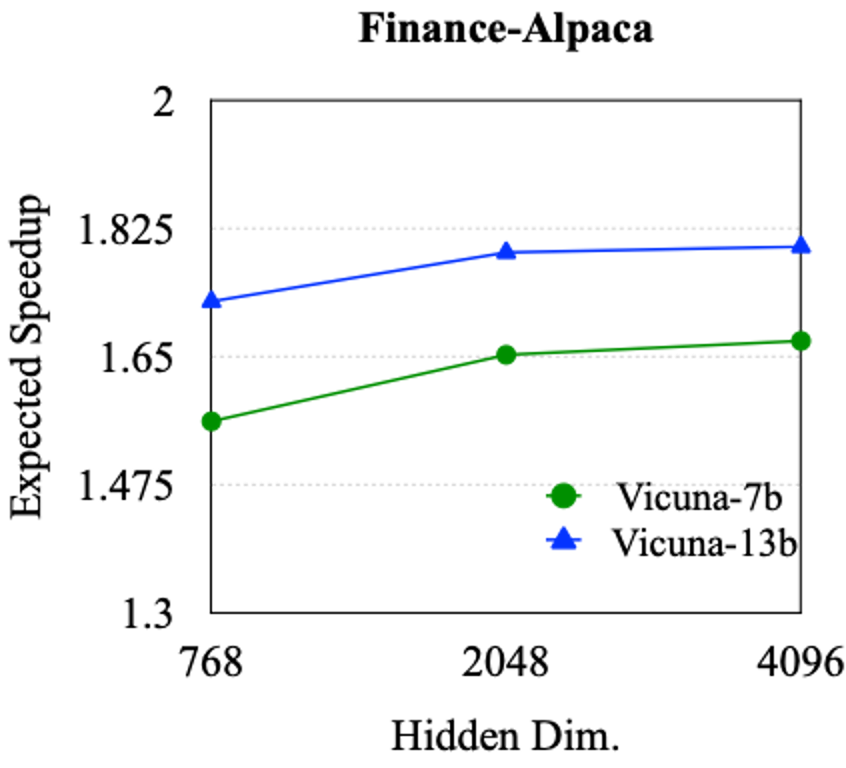}
        \label{fig:sub2}
    \end{minipage}
    \hfill
    \begin{minipage}[b]{0.24\linewidth}
        \centering
        \includegraphics[width=\linewidth]{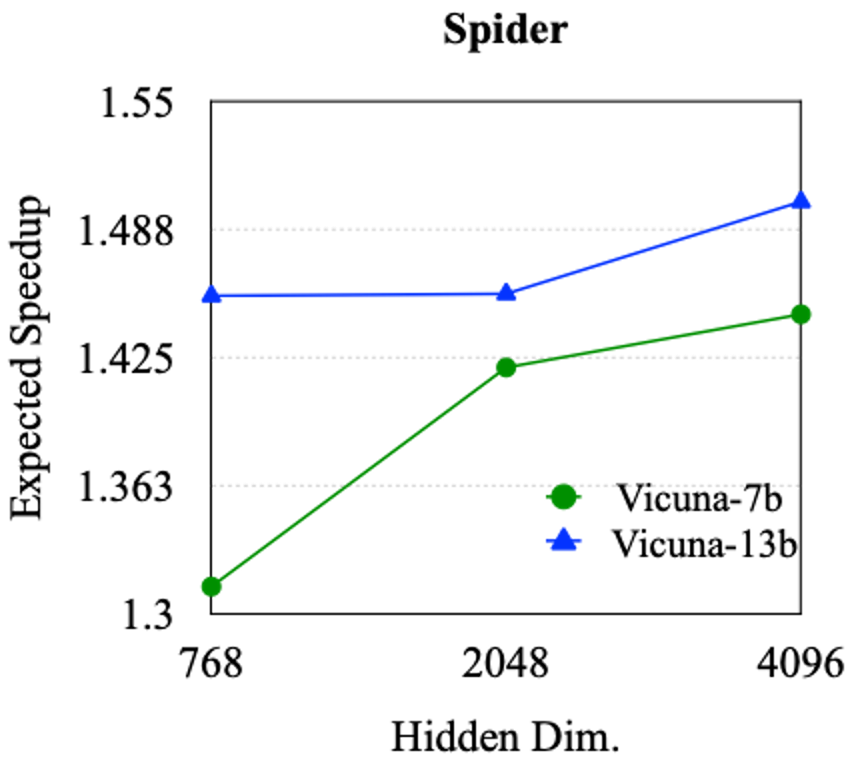}
        \label{fig:sub2}
    \end{minipage}
    \hfill
    \begin{minipage}[b]{0.24\linewidth}
        \centering
        \includegraphics[width=\linewidth]{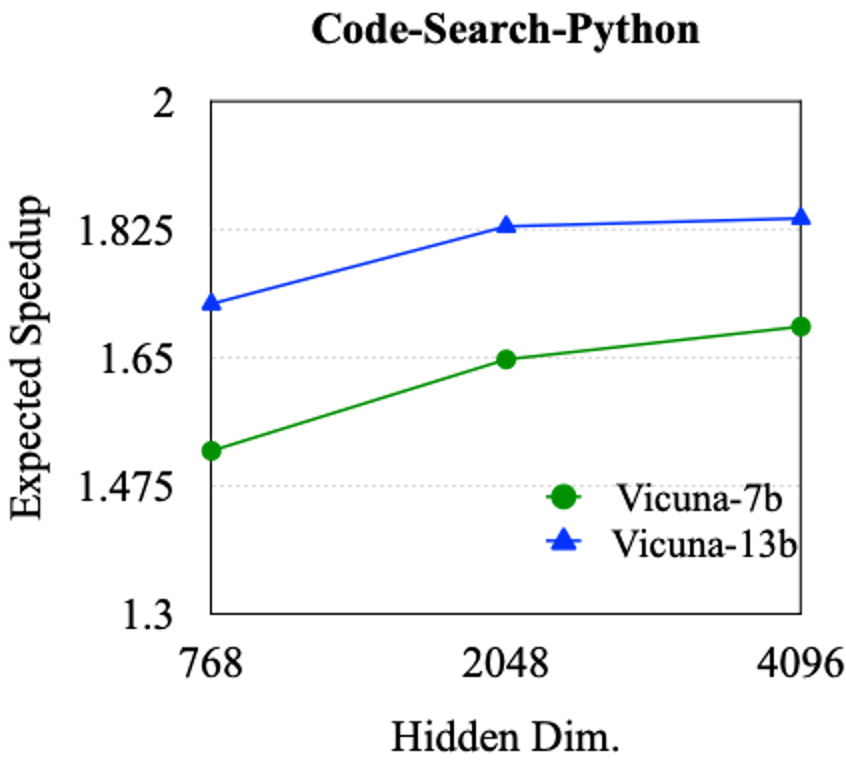}
        \label{fig:sub2}
    \end{minipage}
    \vspace{-0.6cm}
    \caption{The relationship between theoretical speedup and the hidden size of the draft model is such that increasing the hidden size can significantly accelerate the process. This is because the elements within the hidden layer can be computed in parallel, whereas the computation between layers must be done in an autoregressive manner.}
    \label{fig:hidden_scaling}
\end{figure*}
As Table~\ref{tab: apd_batch} shows, if batch sizes are not large enough, it will not largely degenerate the decoding speed. That is because, under the setting of small batch size, the inference process of LLM is memory-bounded.
However, we observe that OOM frequently occurs during for the long prompts, even for the target model only. So we believe how to combine speculative decoding and batch serving systems (e.g., vLLM~\cite{vllm}, continuous batching~\cite{contbach}) will be an important topic in the future.

\section{Configurations for \textsc{GliDe}}
\label{apd: arch}
In our preliminary exploratory research, we discover that the structure of draft models has a significant impact on the speedup ratio. We find that even for the 12-layers llama-160m could lead to a decrease in decoding speed. This is because there is a clear sequential order between layers, necessitating autoregressive operation. Although the multi-layered draft model can significantly improve acceptance rates, this tradeoff is unwise. 

We start with the structure of llama-68m, i.e., 2 hidden layers + hidden dimension=768, and do a grid search to find a more reasonable architecture. 
First, we try settings of 1, 2, and 4 hidden layers (for Vicuna 7B, we only experimented with 1 and 2 layers). We find that for the target models smaller than 33B, the optimal architecture is a single-layer draft model, whereas for the 33B LLM, the optimal architecture is two layers as shown in Fig.~\ref{fig: layer_scaling}. 
Next, we conduct a grid search on the hidden size. We conduct three sets of experiments on the 7B and 13B models, with hidden dimensions of 768, 2048, and 4096, respectively as shown in Figure~\ref{fig:hidden_scaling}. We find that increasing the hidden size can effectively compensate for the loss caused by the reduction of hidden layers, and does not significantly increase operational speed.

\section{Limitation}
Although our model significantly improves acceptance rates, it does have some inherent limitations. The most notable issue is that our model is not a plug-and-play algorithm. This is primarily because our draft model relies on reusing the KV cache of the LLM, necessitating the retraining of a corresponding draft model for speculative decoding each time a new LLM is introduced.  However, considering the relatively low cost of this process, which is approximately 7 hours for training a 68MB draft model on 8 H800 GPU cards, this limitation can be considered manageable within practical applications. We must also note that for vanilla speculative decoding when the tokenizer of the LLM (e.g., Mistral vs Vicuna) is changed, it is necessary to retrain a draft model that is compatible with the new LLM. 
In the training of \textsc{GliDe}, the bottleneck is the forward of the target LLM instead of the lightweight draft model. Therefore, we suggest that future open-source models can be trained simultaneously with a compatible \textsc{GliDe}, thus avoiding the time-consuming re-forward target LLM.
\section{Case Study}
Table~\ref{tab: case1} presents an illustrative case from MT-bench, where tokens marked in red indicate acceptance. The vanilla draft model exhibits a notable deficiency in acceptance rate, evident through a scarcity of red tokens. In contrast, our proposed \textsc{GliDe} demonstrates a markedly higher acceptance of tokens, contributing to increased decoding speed. Notably, \textsc{GliDe} effectively predicts terms such as ``bacteria'' and ``germs'', due to its enhanced drafting capabilities. Furthermore, the integration of \textsc{GliDe} with \textsc{CaPE} shows further improvement in token acceptance, corresponding to the experimental results in Figure \ref{fig: mtbench}. More cases can be found at the following.

\begin{table*}[h]
    \caption{Random Sampled example from MT-bench, \textcolor{red}{red} tokens denotes accepted tokens. The target model is Mistral-7B-instruct-v0.1.}
    \centering
    \begin{tabular}{l m{13.5cm}}
    \toprule

    \bf User Input  & Embrace the role of Sheldon from “The Big Bang Theory” as we delve into our conversation. Don’t start with phrases like “As Sheldon”.  Let’s kick things off with the following question: “What is your opinion on hand dryers?” \\
\midrule
    \bf \textsc{LLaMA}-45m & Hand dryers are a convenient and efficient \textcolor{red}{way to} dry hands after using the restroom or washing them. They can help reduce \textcolor{red}{the} spread \textcolor{red}{of} germ\textcolor{red}{s and} bacteria, as \textcolor{red}{well as} save time \textcolor{red}{and} paper \textcolor{red}{towels}. However\textcolor{red}{,} some people may prefer using paper towels for personal reasons or because they feel more \textcolor{red}{comfortable} with \textcolor{red}{the} sensation. Ul\textcolor{red}{timately}, it comes down \textcolor{red}{to} personal \textcolor{red}{preference} and what works best \textcolor{red}{for} each individual. \\
    \hdashline
    \bf \textsc{GliDe} & Hand dry\textcolor{red}{ers are a} convenient \textcolor{red}{and efficient} way \textcolor{red}{to} dry hands after using the rest\textcolor{red}{room} or \textcolor{red}{washing} them\textcolor{red}{.} They can help \textcolor{red}{reduce the} spread \textcolor{red}{of germs} and \textcolor{red}{bacteria,} as \textcolor{red}{well as} save \textcolor{red}{time and} paper \textcolor{red}{towels.} However\textcolor{red}{,} some people \textcolor{red}{may} prefer using paper \textcolor{red}{towels} for personal \textcolor{red}{reasons} or because \textcolor{red}{they} feel \textcolor{red}{more comfortable} with \textcolor{red}{the} sensation. Ul\textcolor{red}{timately, it} comes \textcolor{red}{down to personal preference} and what works \textcolor{red}{best for} each \textcolor{red}{individual.}\\
    \hdashline
    \bf ~~~+ \textsc{CaPE} & Hand dry\textcolor{red}{ers are a} convenient \textcolor{red}{and efficient} way \textcolor{red}{to} dry \textcolor{red}{hands} after using \textcolor{red}{the} rest\textcolor{red}{room} or \textcolor{red}{washing} them\textcolor{red}{.} \textcolor{red}{They} can \textcolor{red}{help} reduce \textcolor{red}{the spread} of \textcolor{red}{ germs and} \textcolor{red}{bacteria,} \textcolor{red}{as} well \textcolor{red}{as} save \textcolor{red}{time and} paper \textcolor{red}{towels. However}, \textcolor{red}{some} people \textcolor{red}{may prefer} using \textcolor{red}{paper} tow\textcolor{red}{els for} personal \textcolor{red}{reasons or} because \textcolor{red}{they feel} more \textcolor{red}{comfortable} with \textcolor{red}{the} sensation. Ul\textcolor{red}{timately, it} comes \textcolor{red}{down to personal preference} and \textcolor{red}{what} works \textcolor{red}{best for} each \textcolor{red}{individual.}\\
    \bottomrule
    \end{tabular}
    \label{tab: case1}
\end{table*}

\begin{table*}[h]
    \caption{Random Sampled example from MT-bench, \textcolor{red}{red} tokens denotes accepted tokens. Target model is Vicuna-7B-1.5.}
    \centering
    \begin{tabular}{l m{13.5cm}}
    \toprule

    \bf User Input  & Write a descriptive paragraph about a bustling marketplace, incorporating sensory details such as smells, sounds, and visual elements to create an immersive experience for the reader. \\
\midrule
    \bf \textsc{LLaMA}-45m & As I walk \textcolor{red}{through the} b\textcolor{red}{ustling} market\textcolor{red}{place, I} am immediately struck \textcolor{red}{by the} s\textcolor{red}{ights and sounds of} the v\textcolor{red}{ibr}ant atmosphere. The air \textcolor{red}{is} thick with \textcolor{red}{the} s\textcolor{red}{cent of} ex\textcolor{red}{otic} spices \textcolor{red}{and} freshly \textcolor{red}{baked} goods, ming\textcolor{red}{ling} with \textcolor{red}{the} sounds \textcolor{red}{of} vend\textcolor{red}{ors} calling \textcolor{red}{out to} customers and the clatter \textcolor{red}{of} plates and bowls being set out on display\textcolor{red}{.} The market\textcolor{red}{place is a} riot of colors \textcolor{red}{and textures,} with stalls s\textcolor{red}{elling} everything \textcolor{red}{from} brightly hued fab\textcolor{red}{rics to} intric\textcolor{red}{ately} car\textcolor{red}{ved} wooden sculpt\textcolor{red}{ures. The} ground is covered in \textcolor{red}{a} thick \textcolor{red}{layer of} colorful pebbles\textcolor{red}{, and the} walls \textcolor{red}{are} ad\textcolor{red}{orned with} v\textcolor{red}{ibrant} mur\textcolor{red}{als} that seem \textcolor{red}{to} come alive as I walk by\textcolor{red}{.} The market\textcolor{red}{place is a} sens\textcolor{red}{ory} fe\textcolor{red}{ast}, \textcolor{red}{and} I can\textcolor{red}{'t} help \textcolor{red}{but feel} a \textcolor{red}{sense of} excitement and wonder as \textcolor{red}{I} take it all \textcolor{red}{in.} \\
    \hdashline
    \bf \textsc{GliDe} & As I walk \textcolor{red}{through the} b\textcolor{red}{ustling marketplace}, \textcolor{red}{I} am immediately \textcolor{red}{struck by the} s\textcolor{red}{ights and sounds of} the v\textcolor{red}{ibrant} atmosphere\textcolor{red}{. The} air \textcolor{red}{is} thick \textcolor{red}{with the} sc\textcolor{red}{ent of} ex\textcolor{red}{otic spices} and fresh\textcolor{red}{ly baked goods}, mingl\textcolor{red}{ing with the} sounds \textcolor{red}{of} vend\textcolor{red}{ors} calling \textcolor{red}{out to customers} and the cl\textcolor{red}{atter of} pl\textcolor{red}{ates} and bow\textcolor{red}{ls being} set out \textcolor{red}{on} display. \textcolor{red}{The} market\textcolor{red}{place is} a ri\textcolor{red}{ot of} colors and text\textcolor{red}{ures, with} st\textcolor{red}{alls s}elling \textcolor{red}{everything from} bright\textcolor{red}{ly} h\textcolor{red}{ued} fab\textcolor{red}{rics to} intric\textcolor{red}{ately carved wooden} sculpt\textcolor{red}{ures. The} ground is \textcolor{red}{covered in} a thick \textcolor{red}{layer of} color\textcolor{red}{ful} peb\textcolor{red}{bles,} and \textcolor{red}{the} walls \textcolor{red}{are} ad\textcolor{red}{orned with} v\textcolor{red}{ibrant} mur\textcolor{red}{als that} seem \textcolor{red}{to} come \textcolor{red}{alive} as I walk by\textcolor{red}{.} The market\textcolor{red}{place is} a sens\textcolor{red}{ory} fe\textcolor{red}{ast}, \textcolor{red}{and} I \textcolor{red}{can't help} but \textcolor{red}{feel} a \textcolor{red}{sense of excitement and} wonder \textcolor{red}{as I} take it \textcolor{red}{all in.} \\
    \hdashline
    \bf ~~~+ \textsc{CaPE} & As I \textcolor{red}{walk} through \textcolor{red}{the b}ust\textcolor{red}{ling marketplace,} I \textcolor{red}{am} immediately \textcolor{red}{struck by the} s\textcolor{red}{ights and sounds of} the v\textcolor{red}{ibrant atmosphere}. \textcolor{red}{The} air \textcolor{red}{is thick} with \textcolor{red}{the} sc\textcolor{red}{ent of} ex\textcolor{red}{otic spices and} fresh\textcolor{red}{ly baked goods}, m\textcolor{red}{ingling with the} sounds \textcolor{red}{of} vend\textcolor{red}{ors} calling \textcolor{red}{out to customers and} the cl\textcolor{red}{atter of} pl\textcolor{red}{ates and} bow\textcolor{red}{ls being set} out \textcolor{red}{on} display\textcolor{red}{.} The market\textcolor{red}{place is a} ri\textcolor{red}{ot of colors} and \textcolor{red}{text}ures\textcolor{red}{, with} st\textcolor{red}{alls selling} everything \textcolor{red}{from} bright\textcolor{red}{ly} h\textcolor{red}{ued} fab\textcolor{red}{rics to intr}ic\textcolor{red}{ately carved wooden} sculpt\textcolor{red}{ures. The} ground \textcolor{red}{is} covered \textcolor{red}{in a} thick \textcolor{red}{layer of color}ful \textcolor{red}{p}eb\textcolor{red}{bles, and} the walls \textcolor{red}{are ad}orn\textcolor{red}{ed with v}ibr\textcolor{red}{ant mur}als \textcolor{red}{that seem} to \textcolor{red}{come alive} as I \textcolor{red}{walk} by\textcolor{red}{. The} market\textcolor{red}{place is a} sens\textcolor{red}{ory fe}ast\textcolor{red}{,} and \textcolor{red}{I can't} help \textcolor{red}{but feel a sense} of \textcolor{red}{excitement and wonder} as \textcolor{red}{I take} it \textcolor{red}{all in.} \\
    \bottomrule
    \end{tabular}
    \label{tab: case2}
\end{table*}

\begin{table*}[h]
    \caption{Random Sampled example from MT-bench, \textcolor{red}{red} tokens denotes accepted tokens. Target model is Vicuna-7B-1.5.}
    \centering
    \begin{tabular}{l m{13.5cm}}
    \toprule

    \bf User Input  & Picture yourself as a 100-years-old tree in a lush forest, minding your own business, when suddenly, a bunch of deforesters shows up to chop you down. How do you feel when those guys start hacking away at you? \\
\midrule
    \bf \textsc{LLaMA}-45m & As a \textcolor{red}{10}0\textcolor{red}{-year}s-old tree\textcolor{red}{,} I have seen many changes \textcolor{red}{in the} world around \textcolor{red}{me.} I have witness\textcolor{red}{ed} the growth \textcolor{red}{of} for\textcolor{red}{ests} and \textcolor{red}{the} decl\textcolor{red}{ine of} for\textcolor{red}{ests}. \textcolor{red}{I have seen} the rise and \textcolor{red}{fall of} civilizations. \textcolor{red}{I have seen} the impact \textcolor{red}{of} human \textcolor{red}{activity on the} environment.
    \newline\newline
    When the def\textcolor{red}{or}esters show \textcolor{red}{up} and start hack\textcolor{red}{ing} away at me\textcolor{red}{, I} feel a \textcolor{red}{sense of} loss and sad\textcolor{red}{ness. I have} lived a \textcolor{red}{long and} ful\textcolor{red}{filling life}, \textcolor{red}{and} now my time has \textcolor{red}{come to an end}. \textcolor{red}{I} am not afraid of death, \textcolor{red}{but I am} sad \textcolor{red}{to} leave this \textcolor{red}{world} behind\textcolor{red}{.}
    \newline\newline
    \textcolor{red}{I} also feel a \textcolor{red}{sense of} anger \textcolor{red}{and frustration} towards \textcolor{red}{the} def\textcolor{red}{or}esters\textcolor{red}{.} They \textcolor{red}{are} destroy\textcolor{red}{ing the} very thing \textcolor{red}{that} they need \textcolor{red}{to} surv\textcolor{red}{ive} - the for\textcolor{red}{ests} that provide o\textcolor{red}{xygen}, clean water\textcolor{red}{, and} habitat for count\textcolor{red}{less} species\textcolor{red}{.} They \textcolor{red}{are} acting reck\textcolor{red}{lessly and} ir\textcolor{red}{respons}ibly\textcolor{red}{,} without considering \textcolor{red}{the} long\textcolor{red}{-term consequences of} their \textcolor{red}{actions.}
    \newline\newline
    Des\textcolor{red}{pite} my feelings\textcolor{red}{, I} know \textcolor{red}{that I} cannot stop them. I \textcolor{red}{am} just \textcolor{red}{a} tree, after \textcolor{red}{all}. But \textcolor{red}{I} hope \textcolor{red}{that} my sacrifice will serve as \textcolor{red}{a reminder} to future \textcolor{red}{generations} of the importance \textcolor{red}{of} protect\textcolor{red}{ing the} environment and pres\textcolor{red}{erving the} natural world for future \textcolor{red}{generations} to enjoy\textcolor{red}{.} \\
    \hdashline
    \bf \textsc{GliDe} & As \textcolor{red}{a 10}0\textcolor{red}{-year}s\textcolor{red}{-old tree,} I \textcolor{red}{have} seen \textcolor{red}{many changes in} the world around \textcolor{red}{me. I} have witness\textcolor{red}{ed the growth of} for\textcolor{red}{ests} and \textcolor{red}{the} decl\textcolor{red}{ine} of for\textcolor{red}{ests}. \textcolor{red}{I have} seen \textcolor{red}{the} rise and \textcolor{red}{fall of civilizations}. \textcolor{red}{I have seen the} impact \textcolor{red}{of} human activity \textcolor{red}{on the environment.}
    \newline\newline
    When the \textcolor{red}{defor}esters show \textcolor{red}{up} and start hack\textcolor{red}{ing away} at me\textcolor{red}{, I} feel \textcolor{red}{a sense of} loss \textcolor{red}{and} sad\textcolor{red}{ness. I} have lived a \textcolor{red}{long and} ful\textcolor{red}{filling life,} and now my time has \textcolor{red}{come to an end}. \textcolor{red}{I am} not afraid \textcolor{red}{of} death\textcolor{red}{, but I am} sad \textcolor{red}{to} leave this \textcolor{red}{world behind.}
    \newline\newline
    \textcolor{red}{I} also feel \textcolor{red}{a sense of} anger \textcolor{red}{and frustration} towards \textcolor{red}{the} def\textcolor{red}{or}esters\textcolor{red}{.} They are destroy\textcolor{red}{ing the} very thing \textcolor{red}{that} they need \textcolor{red}{to} surv\textcolor{red}{ive} - the for\textcolor{red}{ests that} provide \textcolor{red}{oxygen}, clean water\textcolor{red}{, and} habitat \textcolor{red}{for} count\textcolor{red}{less species.} They \textcolor{red}{are} acting reck\textcolor{red}{lessly and irrespons}ibly\textcolor{red}{,} without considering \textcolor{red}{the} long\textcolor{red}{-term consequences of} their \textcolor{red}{actions.}
    \newline\newline
    Des\textcolor{red}{pite} my \textcolor{red}{feelings, I} know \textcolor{red}{that I} cannot stop \textcolor{red}{them. I am} just \textcolor{red}{a tree,} after \textcolor{red}{all.} But \textcolor{red}{I} hope \textcolor{red}{that} my sacrifice \textcolor{red}{will} serve as \textcolor{red}{a reminder} to future \textcolor{red}{generations of} the \textcolor{red}{importance of protecting} the environment and pres\textcolor{red}{erving the} natural \textcolor{red}{world for future gener}ations to enjoy\textcolor{red}{.} \\
    \hdashline
    \bf ~~~+ \textsc{CaPE} & As \textcolor{red}{a 10}0\textcolor{red}{-year}s\textcolor{red}{-old tree,} I \textcolor{red}{have seen} many \textcolor{red}{changes in} the \textcolor{red}{world} around \textcolor{red}{me. I have} witness\textcolor{red}{ed the growth of} for\textcolor{red}{ests and} the \textcolor{red}{decl}ine \textcolor{red}{of} for\textcolor{red}{ests.} I \textcolor{red}{have seen the} rise \textcolor{red}{and} fall \textcolor{red}{of civilizations.} I \textcolor{red}{have seen the impact} of \textcolor{red}{human} activity \textcolor{red}{on the environment.}
    \newline\newline
    When \textcolor{red}{the} def\textcolor{red}{oresters} show \textcolor{red}{up and} start \textcolor{red}{hack}ing \textcolor{red}{away at me,} I \textcolor{red}{feel} a \textcolor{red}{sense of loss} and \textcolor{red}{sad}ness\textcolor{red}{. I have} lived \textcolor{red}{a} long \textcolor{red}{and ful}fill\textcolor{red}{ing life, and} now my time \textcolor{red}{has} come \textcolor{red}{to an end.} I \textcolor{red}{am not} afraid \textcolor{red}{of} death\textcolor{red}{, but I am} sad \textcolor{red}{to leave} this \textcolor{red}{world behind.}
    \newline\newline
    \textcolor{red}{I also} feel \textcolor{red}{a sense of anger} and \textcolor{red}{frustration towards the} def\textcolor{red}{oresters}. \textcolor{red}{They} are destroy\textcolor{red}{ing the very} thing \textcolor{red}{that} they need \textcolor{red}{to surv}ive \textcolor{red}{-} the for\textcolor{red}{ests that} provide \textcolor{red}{oxygen,} clean \textcolor{red}{water}, \textcolor{red}{and habitat} for count\textcolor{red}{less species. They} are \textcolor{red}{acting reck}lessly \textcolor{red}{and irresponsibly}, \textcolor{red}{without} considering \textcolor{red}{the long}-\textcolor{red}{term consequences of their} actions.
    \newline\newline
    Des\textcolor{red}{pite my} feelings\textcolor{red}{, I} know \textcolor{red}{that I cannot} stop \textcolor{red}{them. I am} just \textcolor{red}{a tree,} after \textcolor{red}{all. But} I \textcolor{red}{hope that my} sacrifice \textcolor{red}{will serve} as \textcolor{red}{a reminder to} future \textcolor{red}{generations of the} importance \textcolor{red}{of protecting the} environment \textcolor{red}{and} pres\textcolor{red}{erving the natural} world \textcolor{red}{for future generations} to \textcolor{red}{enjoy}. \\
    \bottomrule
    \end{tabular}
    \label{tab: case3}
\end{table*}

\end{document}